\documentclass[]{article}
\usepackage{geometry}
\usepackage[affil-it]{authblk}
\usepackage{cite}
\usepackage{amsmath,amssymb,amsfonts}
\usepackage{graphicx}
\usepackage{textcomp}
\usepackage{algorithm}
\usepackage[noend]{algpseudocode}
\usepackage{varwidth}
\usepackage{subcaption}
\usepackage{listings}
\usepackage{color} 
\usepackage[T1]{fontenc}      
\usepackage[framed,numbered]{matlab-prettifier}
\usepackage{verbatim}
\usepackage[hidelinks]{hyperref}
\usepackage{appendix}
\usepackage{wasysym}
\usepackage[breakable, theorems, skins]{tcolorbox}
\hypersetup{
	colorlinks=true,
	linkcolor=blue,
	citecolor=black,
	urlcolor=blue,
}

\definecolor{mygreen}{RGB}{28,172,0} 
\definecolor{mylilas}{RGB}{170,55,241}
\definecolor{mygray}{RGB}{253,249,249}

\newgeometry{vmargin={30mm}, hmargin={30mm,30mm}}

\newtheorem{Ex}{Example}[section]

\title{A Crash Course on Reinforcement Learning}
\author{Farnaz Adib Yaghmaie %
	 \thanks{email: farnaz.adib.yaghmaie@liu.se}}
\affil{Department of Electrical Engineering, Linköping University,\\
	Linköping, Sweden.}

\author{Lennart Ljung%
	\thanks{email: lennart.ljung@liu.se}}
\affil{Department of Electrical Engineering, Linköping University,\\
	Linköping, Sweden.}

\begin{document}

\maketitle

\begin{abstract}
	The emerging field of Reinforcement Learning (RL) has led to impressive results in varied domains like strategy games, robotics, etc. This handout aims to give a simple introduction to RL from control perspective and discuss three possible approaches to solve an RL problem: Policy Gradient, Policy Iteration, and Model-building. Dynamical systems might have discrete action-space like cartpole where two possible actions are +1 and -1 or continuous action space like linear Gaussian systems. Our discussion covers both cases.
\end{abstract}

\section{Introduction}
\textit{Machine Learning (ML)} has surpassed human performance in many challenging tasks like pattern recognition \cite{bishop2006pattern} and playing video games \cite{mnih2013playing}. By recent progress in ML, specifically using deep networks, there is a renewed interest in applying ML techniques to control dynamical systems interacting with a physical environment \cite{duan2016benchmarking,lillicrap2015continuous} to do more demanding tasks like autonomous driving, agile robotics \cite{abbeel2007application}, solving decision-making problems \cite{mnih2015human}, etc. 

Reinforcement Learning (RL) is one of the main branches of Machine Learning which has led to impressive results in varied domains like strategy games, robotics, etc. RL concerned with intelligent decision making in a complex environment in order to maximize some notion of reward. Because of its generality, RL is studied in many disciplines such as control theory \cite{lewis2012reinforcement,lewis2009reinforcement,recht2018tour,matni2019self} and multi-agent systems \cite{Farnaz2017thesis,adib2016output,adib2019differential,adib2018h,yaghmaie2016output,yaghmaie2017bipartite,yaghmaie2015output,yaghmaie2015leader,yaghmaie2016output,yaghmaie2016bipartite,yaghmaie2017multiparty}, etc. RL algorithm have shown impressive performances in many challenging problems including playing Atari games \cite{mnih2013playing}, robotics \cite{abbeel2007application,yaghmaie2012feedback,yaghmaie2013new,yaghmaie2013study}, control of continuous-time systems \cite{Adib2021Linear,Yaghmaie2019using,Yaghmaie2019,Adib2018Output,Adib2018Rein,duan2016benchmarking,lewis2012reinforcement,lewis2009reinforcement,bian2016adaptive,kiumarsi2017h,modares2014linear}, and distributed control of multi-agent systems \cite{Farnaz2017thesis,adib2016output,adib2019differential,yaghmaie2015output}. 

From control theory perspective, a closely related topic to RL is adaptive control theory which studies data-driven approaches for control of unknown dynamical systems \cite{krstic1995nonlinear,aastrom1994adaptive}. If we consider some notion of optimality along with adaptivity, we end up in the RL setting where it is desired to control an unknown system adaptively and optimally. The history of RL dates back decades \cite{sutton1998reinforcement,kaelbling1996reinforcement} but by recent progress in ML, specifically using deep networks, the RL field is also reinvented.

In a typical RL setting, the model of the system is unknown and the aim is to learn how to react with the system to optimize the performance. There are three possible approaches to solve an RL problem \cite{recht2018tour}. 1- \textbf{\textit{Dynamic Programming (DP)}-based solutions}: This approach relies on the principle of optimal control and the celebrated $\mathbf{Q}$\textbf{-learning} \cite{lagoudakis2003least} algorithm is an example of this category. 2- \textbf{Policy Gradient:} The most ambitious method of solving an RL problem is to directly  optimize the performance index \cite{fazel2018global}. 3- \textbf{Model-building RL:} The idea is to estimate a model (possibly recursively) \cite{ljung1999system} and then the optimal control problem is solved for the estimated model. This concept is known as adaptive control \cite{aastrom1994adaptive} in the control community, and there is vast literature around it.

In RL setting, it is important to distinguish between systems with discrete and continuous action spaces. A system with discrete action space has a finite number of actions in each state. An example is the cartpole environment where a pole is attached by an un-actuated joint to a cart \cite{Barto83Neuronlike}. The system is controlled by applying a force of +1 or -1 to the cart. A system with continuous action space has an infinite number of possible actions in each state. Linear quadratic (LQ) control is a well studied example where continuous actions space can be considered \cite{Yaghmaie2019using,Adib2021Linear}. The finiteness or infiniteness of the number of possible actions makes the RL formulation different for these two categories and as such it is not straightforward to use an approach for one to another directly.

In this document, we give a simple introduction to RL from control perspective and discuss three popular approaches to solve RL problems: Policy Gradient, $Q$-learning (as an example of Dynamic Programming-based approach) and model-building method. Our discussion covers both systems with discrete and continuous action spaces while usually the formulation is done for one of these cases. Complementary to this document is a repository called \href{https://github.com/FarnazAdib/Crash_course_on_RL}{A Crash Course on RL}, where one can run the policy gradient and $Q$-learning algorithms on the cartpole and linear quadratic problems.

\subsection{How to use this handout?}
This handout aims to acts as a simple document to explain possible approaches for RL. We do not give expressions and equations in their most exact and elegant mathematical forms. Instead, we try to focus on the main concepts so the equations and expressions may seem sloppy. If you are interested in contributing to the RL field, please consider this handout as a start and deploy exact notation in excellent RL references like \cite{sutton1998reinforcement,bertsekas1995dynamic}.

An important part of understanding RL is the ability to translate concepts to code. In this document, we provide some sample codes (given in shaded areas) to illustrate how a concept/function is coded.  Except for one example in the model-building approach on page \pageref{Shaded:adaptive} which is given in MATLAB syntax (since it uses System Identification toolbox in MATLAB), the coding language in this report is Python. The reason is that Python is currently the most popular programming language in RL. We use \href{https://www.tensorflow.org/}{TensorFlow 2 (TF2)} and \href{https://keras.io/api/}{Keras} for the Machine Learning platforms. TensorFlow 2 is an end-to-end, open-source machine learning platform and Keras is the high-level API of TensorFlow 2: an approchable, highly-productive interface for solving machine learning problems, with a focus on modern deep learning. Keras empowers engineers and researchers to take full advantage of the scalability and cross-platform capabilities of TensorFlow 2. The best reference for understanding the deep learning elements in this handout is \href{https://keras.io/api/}{Keras API reference}.  We use \href{https://gym.openai.com/}{OpenAI Gym} library which is a toolkit for developing and comparing reinforcement learning algorithms \cite{brockman2016openai} in Python. 

The python codes provided in this document are actually parts of a repository called \href{https://github.com/FarnazAdib/Crash_course_on_RL}{A Crash Course on RL}

\begin{center}
	\href{https://github.com/FarnazAdib/Crash_course_on_RL}{https://github.com/FarnazAdib/Crash\_course\_on\_RL}
\end{center}
You can run the codes either in your web browser or in a Python IDE like PyCharm.

\noindent
\textbf{How to run the codes in web browser?} \href{https://jupyter.org/}{Jupyter notebook} is a free and interactive web tool known as a computational notebook, which researchers can use to combine python code and text. One can run Jupyter notebooks (ended with *.ipynb) on Google Colab using web browser. You can run the code by following the steps below: 

\begin{enumerate}
	\item Go to  
	\begin{center}
		\href{https://colab.research.google.com/notebooks/intro.ipynb}{https://colab.research.google.com/notebooks/intro.ipynb}
	\end{center}
	and sign in with a Google account.
	\item Click ``File", and select ``Upload Notebook". If you get the webpage in Swedish, click ``Arkiv" and then ``Ladda upp anteckningsbok".
	\item Then, a window will pop up. Select Github, paste the following link and click search 
	\begin{center}
		\href{https://github.com/FarnazAdib/Crash_course_on_RL}{https://github.com/FarnazAdib/Crash\_course\_on\_RL}
	\end{center}
	\item Then, a list of files with type .ipynb appears. They are Jupyter notebooks. Jupyter notebooks can have both text and code and it is possible to run the code. As an example, scroll down and open ``pg\_on\_cartpole\_notebook.ipynb". 
	\item The file contains some cells with text and come cells with code. The cells which contain code have [ ] on the left. If you move your mouse over [ ], a play box $\RHD$ appears. You can click on it to run the cell. Make sure not to miss a cell as it causes fatal errors.
	\item You can continue like this and run all code cells one by one up to the end.
\end{enumerate}

\noindent
\textbf{How to run the codes in PyCharm?} You can follow these steps to run the code in a Python IDE (preferably PyCharm)
\begin{enumerate}
	\item Go to
	\begin{center}
		\href{https://github.com/FarnazAdib/Crash_course_on_RL}{https://github.com/FarnazAdib/Crash\_course\_on\_RL}
	\end{center}
	and clone the project.
	\item Open PyCharm. From PyCharm. Click File and open project. Then, navigate to the project folder.
	\item Follow \href{https://github.com/FarnazAdib/Crash_course_on_RL/blob/master/Preparation.ipynb}{Preparation.ipynb notebook in ``A Crash Course on RL'' repository}  to build a virtual environment and import required libraries.
	\item Run the python file (ended with .py) you want.
\end{enumerate}

\subsection{Important notes to the reader}
It is important to keep in mind that, the code provided in this document is for illustration purpose; for example, how a concept/function is coded. So do not get lost in Python-related details. Try to focus on how a function is written: what are the inputs? what are the outputs? how this concept is coded? and so on. 

The complete code can be found in \href{https://github.com/FarnazAdib/Crash_course_on_RL}{A Crash Course on RL repository.} The repository contains coding for two classical control problems. The first problem is the cartpole environment which is an example of systems with discrete action space \cite{Barto83Neuronlike}. The second problem is Linear Quadratic problem which is an example of systems with continuous action space \cite{Yaghmaie2019using,Adib2021Linear}. Take the Linear Quadratic problem as a simple example where you can do the mathematical derivations by some simple (but careful) hand-writing. Summaries and simple implementation of the discussed RL algorithms for the cartpole and LQ problem are given in Appendices \ref{App:Cartpole}-\ref{App:LQ}. The appendices are optional, you can skip reading them and study the code directly.

We have summarized the frequently used notations in Table \ref{Table:notation}.

\begin{table*} [!htbp]
	\begin{center}
		\caption{Notation}
		\label{Table:notation}
		\begin{tabular}{p{1.5cm}p{5cm} p{7cm}} 
			\hline
			General: & &\\
			& $[.]^\dagger$ & Transpose operator\\
			&$<\mathcal{S},\mathcal{A},\mathcal{P}, \mathcal{R},\gamma>$ & A Markov Decision Process with state set $\mathcal{S}$, action set $\mathcal{A}$, transition probability set $\mathcal{P}$, immediate reward set $\mathcal{R}$ and discount factor $\gamma$\\
			&$n_s$ & Number of states for discrete state space or dimension of states in continuous action space\\
			&$n_a$ & Number of actions for discrete action space or the dimension of action in continuous action space\\	
			& $\theta$ & The parameter vector to be learned\\		
			& $\pi_{(\theta)}$ & Deterministic policy or probability density function of the policy (with parameter vector $\theta$)\\
			& The subscript $t $ & The time step\\
			& $s_t,\:a_t$ & The state and action at time $t$\\
			& $r_t=r(s_t,a_t)$ & The immediate reward\\
			& $c_t=-r_t$ & The immediate cost\\
			& $R(T)$ & Total reward in form of discounted \eqref{Eq:R:Total}, undiscounted \eqref{Eq:PG:undiscounted_total_reward} or averaged \eqref{Eq:R:Average}.\\
			& $\tau,\:T$ & A trajectory and the trajectory length\\
			Policy Gradient: & & \\
			& $P(\tau|\theta)$ & Probability of trajectory $\tau$ conditioned on $\theta$\\
			&$p(a_t|\theta)$ & evaluation of the parametric pdf $\pi_{\theta}$ at $a_t$ (likelihood)\\
			$Q$-learning: &&\\
			& $V,\:Q$ & The value function and the $Q$-function \\
			&$G$ & The kernel of quadratic $Q=z^{\dagger} G z $\\
			& $\!\begin{aligned}
			\text{vecs}(G)=[&g_{11},...,g_{1n},g_{22},\\
			&...,g_{2n},...,g_{nn}]^{\dagger}
			\end{aligned}$ & The vectorization of the upper-triangular part of a symmetric matrix $G \in \mathbb{R}^{n\times n}$\\
			& $\!\begin{aligned}
			\text{vecv}(v)=[&v_1^2,2v_1 v_2,...,2v_1 v_n,\\
			&v_2^2,...,\: 2v_2 v_n,\: ... ,v_n^2]^{\dagger}
			\end{aligned}$ & The quadratic vector of the vector $v \in \mathbb{R}^{n}$\\
			\hline
		\end{tabular}		
	\end{center}	
\end{table*}

\section{What is Reinforcement Learning}
Machine learning can be divided into three categories: 1- Supervised learning, 2- Unsupervised learning, and 3- Reinforcement Learning (RL). Reinforcement Learning (RL) is concerned with decision making problem. The main thing that makes RL different from supervised and unsupervised learning is that data has a dynamic nature in contrast to static data sets in supervised and unsupervised learning. The dynamic nature of data means that data is generated by a system and the new data depends on the previous actions that the system has received. The most famous definition of RL is given by Sutton and Barto \cite{sutton1998reinforcement} ``Finding suitable actions to take in a given situation in order to maximize a reward".

\begin{figure}	
	\centering	
	\includegraphics[scale=0.4, trim=0cm 0cm 0cm 0cm,clip]{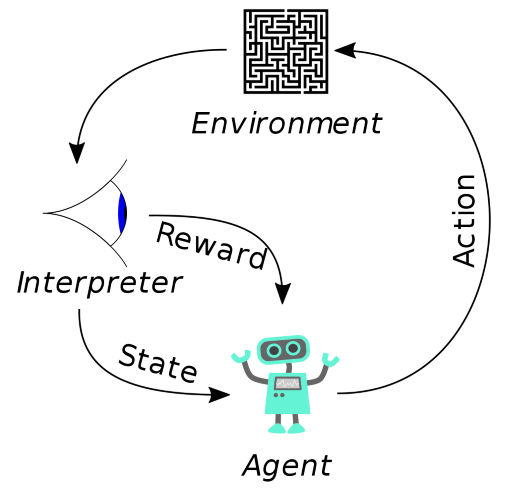}
	\caption{An RL framework. Photo Credit: @ https://en.wikipedia.org/wiki/Reinforcement\_learning}
	\label{Fig:RL:framework}
\end{figure}

The idea can be best described by Fig. \ref{Fig:RL:framework}. We start a loop from the agent. The agent selects an action and applies it to the environment. As a result of this action, the environment changes and reveals a new state, a representation of its internal behavior. The environment reveals a reward which quantifies how good was the action in the given state. The agent receives the state and the reward and tries to select a better action to receive a maximum total of rewards in future. This loop continues forever or the environment reveals a final state, in which the environment will not move anymore.

As we noticed earlier, there are three main components in an RL problem: Environment, reward, and the agent. In the sequel, we introduce these terms briefly.

\subsection{Environment}
Environment is our dynamical system that produces data. Examples of environments are robots, linear and nonlinear dynamical systems (in control theory terminology), and games like Atari and Go. The environment receives an action as the input and generates a variable; namely state; based on its own rules. The rules govern the dynamical model and it is assumed to be unknown. An environment is usually represented by a Markov Decision Process (MDP). In the next section, we will define MDP.

\subsection{Reward}
Along with each state-action pair, the environment reveals a reward $r_t$. Reward is a scalar measurement that shows how good was the action at the state. In RL, we aim to maximize some notion of reward; for example, the total reward where $0 \leq \gamma \leq 1$ is the discount or forgetting factor

\begin{align*}
R= \sum_{t=1}^{T} \gamma^{t}r_t.
\end{align*}

\subsection{Agent}
Agent is what we code. It is the decision-making center that produces the action. The agent receives the state and the reward and produces the action based on some rules. We call such rules \textit{policy} and the agent updates the rules to have a better one. 

\subsubsection{Agent's components}
An RL agent can have up to three main components. Note that the agent need not have all but at least one.

\begin{itemize}
	\item \textbf{Policy:} The policy is the agent's rule to select action in a given state. So, the policy is a map $\pi:\mathcal{S} \rightarrow \mathcal{A}$ from the set of states $\mathcal{S}$ to set of actions $\mathcal{A}$. Though not conceptually correct, it is common to use the terms \textit{``Agent"} and \textit{``Policy"} interchangeably.
	
	\item
	\textbf{Value function}: The value function quantifies the performance of the given policy. It quantifies the expected total reward if we start in a state and always act according to policy.

	\item
	\textbf{Model:} The agent's interpretation of the environment.
\end{itemize}

\subsubsection{Categorizing RL agent}
There are many ways to categorize an RL agent, like model-free and model-based, online or offline agents, and so on. One possible approach is to categorize RL agents based on the main components that the RL agent is built upon. Then, we will have the following classification
\begin{itemize}
	\item Policy gradient.
	\item Dynamic Programming (DP)-based solutions.
	\item Model building.
\end{itemize}
Policy gradient approaches are built upon defining a policy for the agent, DP-based solutions require estimating value functions and model-building approaches try to estimate a model of the environment. This is a coarse classification of approaches; indeed by combining different features of the approaches, we get many useful variations which we do not discuss in this handout.

All aforementioned approaches reduce to some sort of function approximation from data obtained from the dynamical systems. In policy gradient, we fit a function to the policy; i.e. we consider policy as a function of state $\pi=\text{network(state)}$. In DP-based approach, we fit a model to the value function to characterize the cost-to-go. In the model-building approach, we fit a model to the state transition of the environment. 

As you can see, in all approaches, there is a modeling assumption. The thing which makes one approach different from another is ``where'' to put the modeling assumption: policy, value function or dynamical system. The reader should not be confused by the term ``model-free'' and think that no model is built in RL. The term ``model-free'' in RL community is simply used to describe the situation where \textbf{no model of the dynamical system is built}.

\section{Markov Decision Process}
A Markov decision process (MDP) provides a mathematical framework for modeling decision making problems. MDPs are commonly used to describe dynamical systems and represent environment in the RL framework. An MDP is a tuple $<\mathcal{S},\mathcal{A},\mathcal{P}, \mathcal{R},\gamma>$
\begin{itemize}
	\item $\mathcal{S}$: The set of states.
	\item $\mathcal{A}$: The set of actions.
	\item $\mathcal{P}$: The set of transition probability.
	\item $\mathcal{R}$: The set of immediate rewards associated with the state-action pairs.
	\item $0\leq \gamma \leq 1$: Discount factor.
\end{itemize}

%
%

\subsection{States}
It is difficult to define the concept of state but we can say that a state describes the internal status of the MDP. Let $\mathcal{S}$ represent the set of states. If the MDP has a finite number of states, $|\mathcal{S}|=n_s$ denotes the number of states. Otherwise, if the MDP has a continuous action space, $n_s$ denote the dimension of the state vector.

In RL, it is common to define a Boolean variable \textit{done} for each state $s$ visited in the MDP

\begin{align*}
	done(s)=\begin{cases}
	True, \quad \text{$s$ is the final state or the MDP needs to be restarted after $s$}\\
	False, \quad \text{Otherwise.}
	\end{cases}
\end{align*}
This variable is True only if the state is a final state in the MDP: if the MDP goes to this state, the MDP stays there forever or the MDP needs to be restarted. The variable done is False otherwise. Defining done comes handy in developing RL algorithms.

\subsection{Actions}
Actions are possible choices in each state. If there is no choice at all to make, then we have a Markov Process. Let $\mathcal{A}$ represent the set of actions. If the MDP has a finite number of actions, $|\mathcal{A}|=n_a$ denotes the number of actions. Otherwise, if the MDP has a continuous action space, $n_a$ denotes the dimension of the actions. In RL, it is crucial to distinguish between MDPs with discrete or continuous action spaces as the methodology to solve will be different.

\subsection{Transition probability}
The transition probability describes the dynamics of the MDP. It shows the transition probability from all states $s$ to all successor states $s^{\prime}$ for each action $a$. $\mathcal{P}$ is the set of transition probability with $n_a$ matrices each of dimension $n_s \times n_s$ where the $s,\:s^\prime$ entry reads
\begin{align}
[\mathcal{P}^{a}]_{s s^{\prime}}=p[s_{t+1}=s^{\prime} |s_{t}=s,\:a_t=a].
\label{Eq:P:ssa}
\end{align}
One can verify that the row sum is equal to one.

\subsection{Reward}
The immediate reward or reward in short is measure of goodness of action $a_t$ at state $s_t$ and it is represented by 
\begin{align}
r_t=\mathbf{E}[r(s_t,\:a_t)]
\label{Eq:r:sa}
\end{align}
where $t$ is the time index and the expectation is calculated over the possible rewards. $\mathcal{R}$ represent the set of immediate rewards associated with all state-action pairs. In the sequel, we give an example where $r(s_t,\:a_t)$ is stochastic but throughout this handout, we assume that the immediate reward is deterministic and no expectation is involved in \eqref{Eq:r:sa}.

The \textit{total reward} is defined as
\begin{align}
R(T)=\sum_{t=1}^{T} \gamma^{t}r_t,
\label{Eq:R:Total}
\end{align}
where $\gamma$ is the discount factor which will be introduced shortly.

\subsection{Discount factor} 
The discount factor $0\leq \gamma \leq 1$ quantifies how much we care about the immediate rewards and future rewards. We have two extreme cases where $\gamma \rightarrow 0$ and $\gamma \rightarrow 1$. 

\begin{itemize}
	\item $\gamma \rightarrow 0$: We only care about the current reward not what we'll receive in future.
	\item $\gamma \rightarrow 1$: We care all rewards equally.
\end{itemize}

The discounting factor might be given or we might select it ourselves in the RL problem. Usually, we consider $0<\gamma<1$ and more closely to one. We can select $\gamma = 1$ in two cases. 1) There exists an absorbing state in the MDP such that if the MDP is in the absorbing state, it will never move from it. 2) We care about the average cost; i.e. the average of energy consumed in a robotic system. In that case, we can define the average cost as
 \begin{align}
 R(T)=\lim_{T \rightarrow \infty} \frac{1}{T}\sum_{t=1}^{T} r_t.
 \label{Eq:R:Average}
 \end{align}

\begin{Ex}
Consider the MDP in Fig. \ref{Fig:MDP}. This MDP has three states $\mathcal{S}=\{s_0,\: s_1,\:s_2\}$ and two actions $\mathcal{A}=\{a_0,\:a_1\}$. The rewards for some of the transitions are shown by orange arrows. For example, if we start at state $s_1$ and take action $a_0$, we will end up at one of the following cases
\begin{figure}
	\centering
	\includegraphics[scale=0.7, trim=0cm 0cm 0cm 0cm,clip]{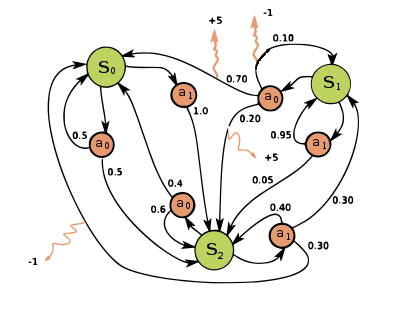}
	\caption{A Markov Decision Process. The photo is a modified version of the photo in @ https://en.wikipedia.org/ wiki/Markov\_decision\_process}
	\label{Fig:MDP}
\end{figure}
\begin{itemize}
	\item With probability $0.1$, the reward is $-1$ and the next state is $s_1$.
	\item With probability $0.7$, the reward is $+5$ and the next state is $s_0$.
	\item With probability $0.2$, the reward is $+5$ and the next state is $s_2$
\end{itemize}
As a result, the reward for state $s_1$ and action $a_0$ reads
\begin{align*}
\mathbf{E}[r(s_1,\:a_0)]=0.1 \times (-1)+0.7 \times (5) + + 0.2 \times(5) = 4.4.
\end{align*}

The transition probability matrices are given by
\begin{align*}
\mathcal{P}^{a_0}=\begin{bmatrix}
0.5 & 0   & 0.5\\
0.7 & 0.1 & 0.2\\
0.4 & 0   & 0.6
\end{bmatrix},\: 
\mathcal{P}^{a_1}=\begin{bmatrix}
0   & 0   & 1\\
0   & 0.95& 0.05\\
0.3 & 0.3 & 0.4
\end{bmatrix}.
\end{align*}
Observe that the sum of each row in $\mathcal{P}^{a_0},\:\mathcal{P}^{a_1}$ equals to one. 
\end{Ex}
	
\subsection{Revisiting the agents component again}
Now that we have defined MDP, we can revisit the agents components and define them better. As we mentioned an RL agent can have up to three main components. 
\begin{itemize}
	\item \textbf{Policy:} The policy is the agent's rule to select action in a given state. So, the policy is a map $\pi:\mathcal{S} \rightarrow \mathcal{A}$. We can have Deterministic policy $a=\pi(s)$ or stochastic policy defined by a pdf $\pi(a|s)=P\left[ a_t=a | s_t=s\right]   $.
	
	\item
	\textbf{Value function}: The value function quantifies the performance of the given policy in the states
	\begin{align*}
	V(s)=\mathbf{E}\left[ r_{t} + \gamma r_{t+1}+\gamma^2 r_{t+2} +... | s_t=s\right].
	\end{align*}
	
	\item
	\textbf{Model:} The agent's interpretation of the environment $[\mathcal{P}^{a}]_{s s^{\prime}}$ which might be different from the true value.
\end{itemize}
We categorize possible approaches to solve an RL problem based on the main component on which the agent is built upon. We start with the policy gradient approach in the next section which relies on building/estimating policy.
\section{Policy Gradient}
\label{Sec:PG}
The most ambitious method of solving an RL problem is to directly learn the policy from optimizing the total reward. We do not build a model of environment and we do not appeal to the Bellman equation. Indeed our modeling assumption is in considering a parametric probability density function for the policy and we aim to learn the parameter to maximize the expected total reward

\begin{align}
\large
J = \mathbf{E}_{\tau \sim \pi_{\theta}}[R(T)]
\label{Eq:PG:J}
\end{align}
where
\begin{itemize}
	\item $\large \pi_{\theta}$ is the probability density function (pdf) of the policy and $\theta$ is the parameter vector. 
	\item $\large \tau$ is a trajectory obtained from sampling the policy and it is given by 
	\begin{align*}
		\tau =(s_{1},\:a_{1},\:r_{1},\: s_{2},\: a_{2},\:r_{2},\:s_{3},...,s_{T+1})
	\end{align*}
	where $s_{t},\:a_{t},\:r_{t}$ are the state, action, reward at time $t$ and $T$ is the trajectory length. $\tau \sim \pi_{\theta}$ means that trajectory $\tau$ is generated by sampling actions from the pdf $\pi_{\theta}$.
	\item $R(T)$ is undiscounted finite-time total reward
	\begin{align}
	R(T)= \sum_{t=1}^{T} r_t.\: \label{Eq:PG:undiscounted_total_reward}
	\end{align}
	\item Expectation is defined over the probability of the trajectory
\end{itemize}
We would like to directly optimize the policy by a gradient approach. So, we aim to obtain the gradient of $J$ with respect to parameter $\theta$

\begin{align*}
\large
\nabla_{\theta} J.
\end{align*}

The algorithms that optimizes the policy in this way are called\textit{ Policy Gradient (PG)} algorithms. The log-derivative trick helps us to obtain the policy gradient $\large \nabla_{\theta} J$. The trick depends on the simple math rule $\nabla_{p} \log p=\dfrac{1}{p}$. Assume that $p$ is a function of $\theta$. Then, using chain rule, we have

\begin{align*}
\large
\nabla_{\theta} \log p = \nabla_{p} \log p \nabla_{\theta} p =  \dfrac{1}{p}\nabla_{\theta} p .
\end{align*}
Rearranging the above equation
\begin{align}
\large
\nabla_{\theta} p =p \nabla_{\theta} \log p.
\label{Eq:PG:Log:derivative}
\end{align}
Equation \eqref{Eq:PG:Log:derivative} is called the log-derivative trick and helps us to get rid of dynamics in PG. You will see an application of \eqref{Eq:PG:Log:derivative} in Subsection \ref{Subsec:PG:gradient}.

In the sequel, we define the main components in PG.

\subsection{Defining probability density function for the policy}
\label{Subsec:PG:pdf}
In PG, we consider the class of stochastic policies. One may ask why do we consider stochastic policies when we know that the optimal policy for MDP is deterministic \cite{puterman2014markov,recht2018tour}? The reason is that in PG, no value function and no model of the dynamics are built. The only way to evaluate a policy is to deviate from it and see the total reward.  So, the burden of the optimization is shifted onto sampling the policy: By perturbing the policy and observing the result, we can improve policy parameters.
If we consider a deterministic policy in PG, the agent gets trapped in a local minimum. The reason is that the agent has ``no'' way of examining other possible actions and furthermore, there is no value function to show how ``good'' the current policy is. Considering a stochastic policy is essential in PG.

\begin{figure}[h]
	\centering
	\includegraphics[scale=0.35, trim=0cm 0cm 0cm 0cm,clip]{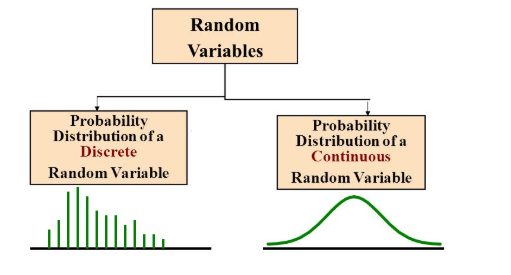}
	\caption{Pdf for discrete and continuous reandom variables. Photo Credit: @ https://towardsdatascience.com/probability-distributions-discrete-and-continuous-7a94ede66dc0}
	\label{Fig:PG:random}
\end{figure}

As a result, our modeling assumption in PG is in considering a probability density function (pdf) for the policy. As we can see in Fig. \ref{Fig:PG:random} the pdf is defined differently for discrete and continuous random variables. For discrete random variables, the pdf is given as probability for all possible outcomes while for continuous random variables it is given as a function. This tiny technical point makes coding completely different for the discrete and continuous action space cases. So we treat discrete and continuous action spaces differently in the sequel.

\subsubsection{Discrete action space}
\label{Subsubsec:PG:pdf:Discrete}
As we said earlier, our modeling assumption in PG is in considering a parametric pdf for the policy. We represent the pdf with $\pi_{\theta}$ where $\theta$ is the parameter. The pdf $\pi_{\theta}$ maps from the state to the probability of each action. So, if there are $n_a$ actions, the policy network has $n_a$ outputs, each representing the probability of an action. Note that the outputs should sum to 1. 

\begin{figure}	
	\centering	
	\includegraphics[scale=0.4, trim=0cm 0cm 0cm 0cm,clip]{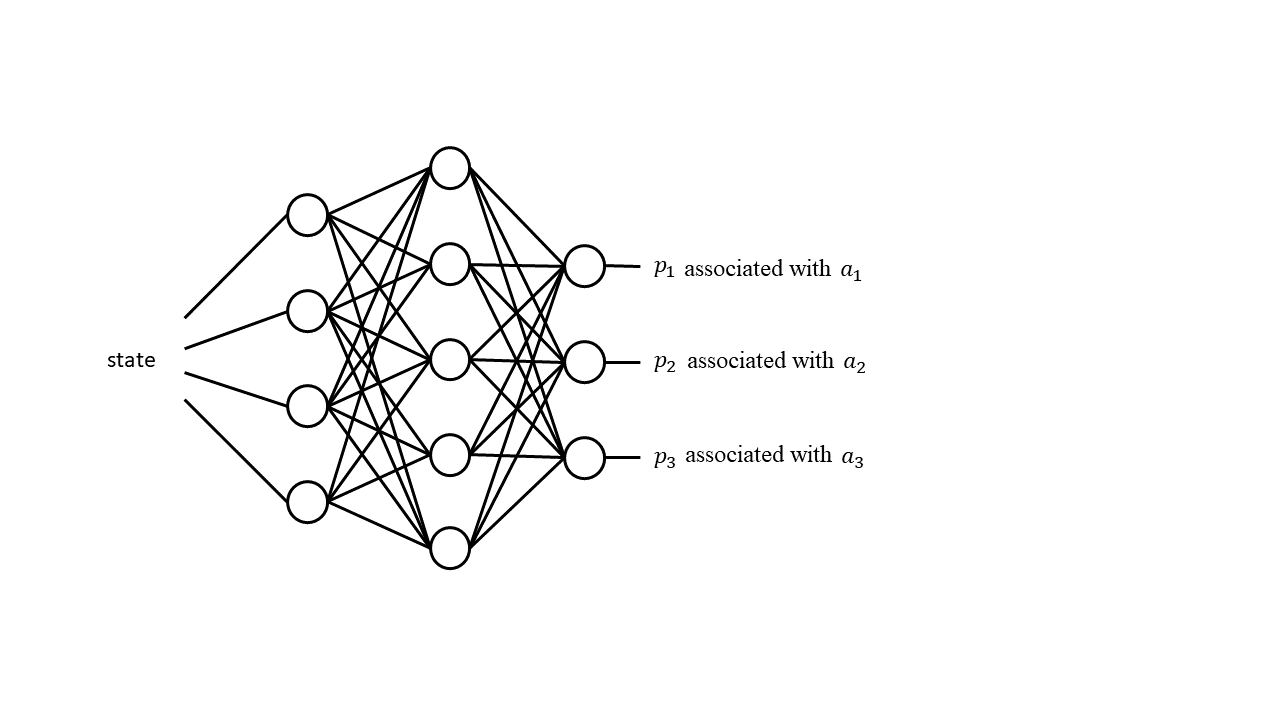}
	\caption{An example of network producing the pdf $\pi_{\theta}$}
	\label{Fig:PG:pi}
\end{figure}

An example of network is shown in Fig. \ref{Fig:PG:pi}. The network generates the pdf for three possible actions by taking state as the input. In this figure, $p_1$ is the probability associated with action $a_1$, $p_2$ associated with action $a_2$ and $p_3$ is associated with action $a_3$. Note that it should hold $p_1+p_2+p_3=1$. 

\begin{tcolorbox}[left skip=0.5cm, before skip=0.5cm,
	breakable,arc=8pt,borderline={0pt}{0pt}{white},boxrule=0mm,fontupper=\small]
	\textit{\textbf{Generating pdf and sampling an action in discrete action space case}}
	
Let $\pi_{\theta}$ be generated by the function \lstinline[language=Python]!network(state)!

\begin{lstlisting}[language=Python]
network = keras.Sequential([
keras.layers.Dense(30, input_dim=n_s, activation='relu'),
keras.layers.Dense(30, activation='relu'),
keras.layers.Dense(n_a, activation='softmax')])
\end{lstlisting}

In the above code, the network is built and the parameters of the network (which are biases and weights) are initialized. The network takes state of dimension $n_s$ as the input and uses it in a fully connected layer with 30 neurons, with the activation function as relu, followed by another layer with 30 neurons and again with the activation function as relu. Then, we have the last layer which has $n_a$ number of outputs and we select the activation function as softmax as we want to have the sum of probability equal to one. 

To draw a sample $a \sim \pi_{\theta}$, first we feed the state to the network to produce the pdf $\pi_{\theta}$ and then, we select an action according to the pdf. This can be done by the following lines of code
\begin{lstlisting}[language=Python]
softmax_out = network(state)
a = np.random.choice(n_a, p=softmax_out.numpy()[0])
\end{lstlisting}
\end{tcolorbox}

\subsubsection{Continuous action space}
\label{Subsubsec:PG:pdf:Continuous}
When the action space is continuous, we select the pdf $\pi_{\theta}$ as a diagonal Gaussian distribution $ \pi_{\theta}=\mathcal{N}(\mu_{\theta},\Sigma)$, where the mean is parametric and the covariance is selected as $\Sigma= \sigma^2 I_{n_a}$, with $\sigma>0$ as a design parameter

\begin{align*}
\pi_{\theta}= \dfrac{1}{\sqrt{(2 \pi \sigma^2)^{n_{a}}}} \exp[-\dfrac{1}{2\sigma^2} (a-\mu_{\theta}(s))^{\dagger}(a-\mu_{\theta}(s))].
\end{align*}
As a result, our modeling assumption is in the mean of the pdf, the part that builds our policy $\mu_{\theta}$. The actions are then sampled from the pdf $\pi_{\theta}=\mathcal{N}(\mu_{\theta},\Sigma)$. For example, a linear policy can be represented by $\mu_{\theta} =\theta s$ where $\theta$ is the linear gain and the actions are sampled from $\mathcal{N}(\theta s,\sigma^2 I_{n_a})$.

\begin{tcolorbox}[left skip=0.5cm, before skip=0.5cm,
	breakable,arc=8pt,borderline={0pt}{0pt}{white},boxrule=0mm,fontupper=\small]
	\textit{\textbf{Sampling an action in continuous action space}}
	
Let $\mu_{\theta}$ be generated by the function \lstinline[language=Python]!network(state)!. That is $\mu_{\theta}(s)=$ \lstinline[language=Python]!network(state)! takes the state variable as the input and has vector parameter $\theta$. To draw a sample $a \sim \mathcal{N}(\mu_{\theta},\sigma I_{n_a})$, we do the following

\begin{lstlisting}[language=Python]
a = network (state) + sigma * np.random.randn(n_a)
\end{lstlisting}
\end{tcolorbox}

\subsection{Defining the probability of trajectory}
\label{Subsec:Prob:Traj}
We defined a parametric pdf for the policy in the previous subsection. The next step is to sample actions from the pdf and generate a trajectory. $\tau \sim \pi_{\theta}$ means that a trajectory of the environment is generated by sampling action from $\pi_{\theta}$. Let $s_{1}$ denote the initial state of the environment. The procedure is as follows. 
\begin{enumerate}
	\item We sample the action $a_{1}$ from the pdf; i.e. $a_{1 }\sim \pi_{\theta}$. We derive the environment using $a_{1}$. The environment reveals the reward $r_{1}$ and transits to a new state $s_{2}$.
	\item  We sample the action $a_{2}$ from the pdf; i.e. $a_{2} \sim \pi_{\theta}$. We derive the environment using $a_{2}$. The environment reveals the reward $r_{2}$ and transits to a new state $s_{3}$. 
	\item We repeat step 2 for $T$ times and in the end, we get a trajectory 
	\begin{align*}
		\tau =(s_{1},\:a_{1},\:r_{1},\: s_{2},\: a_{2},\:r_{2},\:s_{3},...,s_{T+1}).
	\end{align*}
	
\end{enumerate}

The probability of the trajectory $\tau$ is defined as follows

\begin{align}
\large
P(\tau| \theta) = \prod_{t=1}^{T}p(s_{t+1}|s_{t},a_{t}) p(a_t|\theta).
\label{Eq:PG:probability:trajectory}
\end{align}
in which
\begin{itemize}
	\item $p(s_{t+1}|s_{t},a_{t})$ represents the dynamics of the environment; it defines the next state $s_{t+1}$ given the current state $s_{t}$ and the current action $a_{t}$. Note that in RL we do NOT know $p(s_{t+1}|s_{t},a_{t})$. You will see later that $p(s_{t+1}|s_{t},a_{t})$ is not needed in the computation.
	
	\item $p(a_{t}|\theta)$ is the likelihood function and it is obtained by evaluating the pdf $\pi_{\theta}$ at $a_{t}$. In the sequel, we will see how $p(a_{t}|\theta)$ is defined in discrete and continuous action spaces.
\end{itemize}

\subsubsection{Discrete action space}
If the action space is discrete, \lstinline[language=Python]!network(state)! denotes the probability density function $\pi_{\theta}$. It is a vector with however many entries as there are actions, and the actions are the indices for the vector. So, $p(a_{t}|\theta)$ is obtained by indexing into the output vector \lstinline[language=Python]!network(state)!.

\subsubsection{Continuous action space}
\label{Subsubsec:likelihood:continuous}
Let the action space be continuous and assume that the dimension is $n_a$, we consider a multi-variate Gaussian with mean $\mu_{\theta}(s)=$\lstinline[language=Python]!network(state)!. Then, $p(a_t|\theta)$ is given by

\begin{align}
p(a_t|\theta) = \dfrac{1}{\sqrt{(2 \pi \sigma^2)^{n_a}}} \exp[-\dfrac{1}{2\sigma^2} (a_t-\mu_{\theta}(s_t))^{\dagger}(a_t-\mu_{\theta}(s_t))].
\label{Eq:PG:Trajectory:Continuous}
\end{align}

\subsection{Computing the gradient $\nabla_{\theta} J$}
\label{Subsec:PG:gradient}
The final step in PG which results in learning the parameter vector is to compute the gradient of $J$ in \eqref{Eq:PG:J}-\eqref{Eq:PG:undiscounted_total_reward} with respect to the parameter vector $\theta$; that is $\nabla_{\theta} J$. We already have all components to compute this term. First, we need to do a little math here 

\begin{align}
\begin{split}
\nabla_{\theta} J&=\nabla_{\theta} \mathbf{E}\: [ R(T)]\\
&= \nabla_{\theta} \int_{\tau} P(\tau| \theta) R(T) \quad \text{replacing the expectation with the integral,}\\
&= \int_{\tau} \nabla_{\theta} P(\tau| \theta) R(T) \quad \text{bringing the derivative inside,}\\
&= \int_{\tau} P(\tau| \theta) \nabla_{\theta} \log P(\tau| \theta) R(T)\quad \text{using log-derivative trick \eqref{Eq:PG:Log:derivative},}\\
&= \mathbf{E} [\nabla_{\theta} \log P(\tau| \theta) R(T)] \quad \text{replacing the integral with the expectation.}
\end{split}
\label{Eq:PG:gradient:derivation}
\end{align}
In \eqref{Eq:PG:gradient:derivation}, $P(\tau| \theta)$  is the probability of the trajectory defined in \eqref{Eq:PG:Log:derivative}. $\nabla_{\theta} \log P(\tau| \theta)$ reads

\begin{align}
\begin{split}
\nabla_{\theta} \log P(\tau| \theta)&=\nabla_{\theta} \sum_{t=1}^{T} \log p(s_{t+1}|s_{t},a_{t}) +\nabla_{\theta}\sum_{t=1}^{T} \log p(a_{t}|\theta)\\
&=\sum_{t=1}^{T}\nabla_{\theta} \log p(a_{t}|\theta).
\end{split}
\label{Eq:PG:gradient:trajectory}
\end{align}
The first summation in \eqref{Eq:PG:gradient:trajectory} contains the dynamics of the system $\log p(s_{t+1}|s_{t},a_{t})$ but since it is independent of $\theta$, it disappears while taking gradient. $p(a_{t}|\theta)$ is the likelihood function defined in subsection \ref{Subsec:Prob:Traj} for continuous (see \eqref{Eq:PG:Trajectory:Continuous}) and discrete action spaces. By substituting \eqref{Eq:PG:gradient:trajectory} in \eqref{Eq:PG:gradient:derivation} $\nabla_{\theta} J$ reads

\begin{align}
\large
\nabla_{\theta} J= \mathbf{E} [R(T)\sum_{t=1}^{T}\nabla_{\theta} \log p(a_{t}|\theta)]. 
\label{Eq:PG:gradient}
\end{align}
This is the main equation in PG.  One can replace the expectation with averaging or simply drop the expectation operator.

\subsubsection{Discrete action space}
\label{Subsubsec:PG:Gradient:Discrete}
Computing \eqref{Eq:PG:gradient} in the discrete action space case is quite simple because we can use a pre-built cost function in Machine learning libraries. To see this point note that $J$ (without the gradient)
\begin{align}
\large
J= \sum_{t=1}^{T}R(T)\log p(a_{t}|\theta)
\label{Eq:PG:performance}
\end{align}
is in the form of the weighted cross entropy cost (wcec) function which is used and optimized in the classification task
\begin{align}
	J_{wcec} = -\frac{1}{M} \sum_{m=1}^{M} \sum_{c=1}^{C}  w_{c} \times y_{m}^{c} \times \log(h_{\theta}(x_{m}, c))
	\label{Eq:PG:wcec}
\end{align}
where
\begin{itemize}
	\item $C$: number of classes,
	\item $M$: number of training data,
	\item $w_c$: is the weight of class $c$,
	\item $x_m$: input for training example $m$,
	\item $y_{m}^{c}$: target label for $x_m$ for class $c$,
	
	\item $h_{\theta}$: neural network producing probability with parameters $\theta$.
\end{itemize}

At the first glance, it might seem difficult to recast the performance index \eqref{Eq:PG:performance} to the weighted cross entropy cost function in \eqref{Eq:PG:wcec}. But a closer look will verify that it is indeed possible. We aim to maximize \eqref{Eq:PG:performance} in PG while in the classification task, the aim is to minimize the weighted cross entropy cost in \eqref{Eq:PG:wcec}. This resolves the minus sign in \eqref{Eq:PG:wcec}. 
$n_a$ actions are analogous to $C$ categories and the trajectory length $T$ in \eqref{Eq:PG:performance} is analogous to the number of data $M$ in \eqref{Eq:PG:wcec}.  $R(T)$ is the weight of class $c$; i.e. $w_c$. $x_m$ is analogous to the state $s_t$. $y_{m}^{c}$ is the target label for training example $m$ for class $c$,
\begin{align*}
	y_{m}^{c}=\begin{cases}
	1 \quad \text{if $c$ is the correct class for $x_m$},\\
	0 \quad \text{otherwise.}
	\end{cases}
\end{align*}
In \eqref{Eq:PG:performance}, the target label is defined similarly and hides the summation over actions. That is, we label data in the following sense. Assume that at state $s_t$, the action $a_t$ is sampled from the pdf. Then, the target label for state $s_t$ and action $a$ is defined as follows:
\begin{align*}
y_{t}^{a}=\begin{cases}
1 \quad \text{if $a=a_t$},\\
0 \quad \text{otherwise.}
\end{cases}
\end{align*}
Finally $h_{\theta}(x_{m}, k)$ is analogous to the probability of the selected action $a_t$ which can be obtained from the output of the network for the state $s_t$.

In summary, we can optimize $J$ in \eqref{Eq:PG:performance} in a similar way that the cost function in the classification task is minimized. To do so, we need to recast our problem to a classification task, meaning that our network should produce probability in the last layer, we need to label data, and define the cost to be optimized as the weighted cross entropy.

\begin{tcolorbox}[left skip=0.5cm, before skip=0.5cm,
	breakable,arc=8pt,borderline={0pt}{0pt}{white},boxrule=0mm,fontupper=\small]
	\textit{\textbf{Learning parameter in discrete action space case}}
	
Let \lstinline[language=Python]!network(state)! represent the parametric pdf of the policy in the discrete action space case. We define a cross entropy loss function for the network
\begin{lstlisting}[language=Python]
network.compile(loss='categorical_crossentropy')
\end{lstlisting}
Now, we have configured the network and all we need to do is to pass data to our network in the learning loop. To cast \eqref{Eq:PG:gradient} to the cost function in the classification task, we need to define the true probability for the selected action. In other words, we need to label data. For example, if we have three different actions and the second action is sampled, the true probability or the labeled data is $[0,1,0]$. The following line of the code, produces labeled data based on the selected action
\begin{lstlisting}[language=Python]
target_action = tf.keras.utils.to_categorical(action, n_a)
\end{lstlisting}
Now, we compute the loss of the network by giving the state, the target\_action, and weighting $R(T)$. The \lstinline[language=Python]!network(state)! gets the state as the input and creates the probability density functions in the output. The true probability density function is defined by \lstinline[language=Python]!target_action! and it is weighted by \lstinline[language=Python]!R_T!. That is it!
\begin{lstlisting}[language=Python]
loss = network.train_on_batch(state, target_action, 
sample_weight=R_T)
\end{lstlisting}
\end{tcolorbox}

\subsubsection{Continuous action space}
\label{Subsubsec:PG:Gradient:Continuous}
Remember that for continuous action space, we have chosen a multi-variate Gaussian distribution for the pdf, see subsections \ref{Subsubsec:PG:pdf:Continuous} and \ref{Subsubsec:likelihood:continuous}. Based on \eqref{Eq:PG:Trajectory:Continuous}, we have 
\begin{align}
	\nabla_{\theta} \log p(a_{t}|\theta)= \frac{1}{\sigma^2}\frac{d \mu_{\theta}(s_{t})}{d \theta}(a_{t}-\mu_{\theta}(s_{t})).
	\label{Eq:PG:gradient_likelihood}
\end{align}
To evaluate the gradient, we sample $\mathcal{D}$ trajectories and replace the expectation with the average of $|\mathcal{D}|$ trajectories. Then, using \eqref{Eq:PG:gradient_likelihood} $\nabla_{\theta} J$ in \eqref{Eq:PG:gradient} reads

\begin{align}
\large
\nabla_{\theta} J = \dfrac{1}{\sigma^2 |\mathcal{D}|}\sum_{\tau \in \mathcal{D}} \sum_{t=1}^{T}(a_{t}-\mu_{\theta}(s_{t}))\frac{d \mu_{\theta}(s_{t})}{d \theta}^{\dagger} R(T).
\label{Eq:PG:gradient:continuous}
\end{align}
For example, if we consider a linear policy $\mu_{\theta}(s_{t})= \theta \: s_{t}$, \eqref{Eq:PG:gradient:continuous} is simplified to

\begin{align}
\large
\nabla_{\theta} J = \dfrac{1}{\sigma^2 |\mathcal{D}|}\sum_{\tau \in \mathcal{D}} \sum_{t=1}^{T}(a_{t}-\theta \:s_{t}) s_t^{\dagger} R(T).
\label{Eq:PG:gradient:continuous:linear}
\end{align}
Then, we can improve the policy parameter $\theta$ by a gradient approach.

\subsection{PG as an Algorithm}
First, we build/consider a parametric pdf $\pi_\theta(s)$, see subsection \ref{Subsec:PG:pdf}. Then, we iteratively update the parameter $\theta$. In each iteration of the algorithm, we do the following
\begin{enumerate}
	\item We sample a trajectory from the environment to collect data for PG by following these steps:
	\begin{enumerate}
		\item We initialize empty histories for \textit{states=[], actions=[], rewards=[]}.
		\item We observe the state $s$ and sample action $a$ from the policy pdf $\pi_{\theta}(s)$. See subsection \ref{Subsec:PG:pdf}.
		\item We derive the environment using $a$ and observe the reward $r$.
		\item We add $s,\:a,\:r$ to the history batch \textit{states, actions, rewards}.
		\item We continue from 1.(b) until the episode ends.
	\end{enumerate}
	\item We improve the policy by following these steps:
	\begin{enumerate}
		\item We calculate the total reward \eqref{Eq:PG:undiscounted_total_reward}.
		\item We optimize the parameters policy. See subsection \ref{Subsec:PG:gradient}. 
	\end{enumerate}
\end{enumerate}

\subsection{Improving PG}
While PG is an elegant algorithm, it does not always produce good (or any) result . There are many approaches that one can use to improve the performance of PG. The first approach is to consider ``reward-to-go" 
\begin{align}
	R_{T}(t)= \sum_{k=t}^{T}r_k.
	\label{Eq:PG:R_to_go}
\end{align}
instead of total reward \eqref{Eq:PG:undiscounted_total_reward}. The reason is that the rewards obtained before time $t$ is not relevant to the state and action at time $t$. The gradient then reads
\begin{align}
\large
\nabla_{\theta} J= \mathbf{E}_{\tau \sim \pi_{\theta}} [\sum_{t=1}^{T}R_T(t)\nabla_{\theta} \log p(a_{t}|\theta)]. 
\label{Eq:PG:gradient:R_to_go}
\end{align}

Another possible approach is to subtract a baseline $b$ from the total cost \eqref{Eq:PG:undiscounted_total_reward} or the cost-to-go. The gradient then reads
\begin{align}
\large
\nabla_{\theta} J= \mathbf{E}_{\tau \sim \pi_{\theta}} [\sum_{t=1}^{T}(R_T(t)-b)\nabla_{\theta} \log p(a_{t}|\theta)]. 
\label{Eq:PG:gradient:baseline}
\end{align}
The justification is that if we subtract a constant from the objective function in an optimization problem, the minimizing argument does not change. Subtracting baseline in PG acts as a standardization of the optimal problem and can accelerate computation. See \cite{matni2019self} for possible choices for the baseline function.

There are other possible approaches in the literature to improve PG that we have not discussed here. Note that not all of these methods improve the performance of PG for a specific problem and one should carefully study the effect of these approaches and select the one which works.

\section{$Q$ learning}
\label{Sec:Q:learing}
Another possible approach to solve an RL problem is to use Dynamic Programming (DP) and assort to Bellman's principle of optimality. Such approaches are called Dynamic-Programming based solutions. The most popular DP approach is $Q$ learning which relies on the definition of quality function. Note that in $Q$ learning, we parameterize the quality function and the policy is defined by maximizing (or minimizing depending on whether you consider reward or cost) the $Q$-function. In $Q$ learning our modeling assumption is in considering a parametric structure for the $Q$ function. 
%

\subsection{$Q$ function}
\label{Subsec:Q}
The $Q$ function is equal to the expected reward for taking an arbitrary action $a$ and then following the policy $\pi$. In this sense, the $Q$ function quantifies the performance of a policy in each state-action pair 

\begin{align}
\large
Q(s,a) =r(s,a)+ \gamma\: \mathbf{E}[Q(s^{\prime}, \pi(s^{\prime}))]
\label{Eq:Q}
\end{align}
where the policy $\large \pi$ is the action maximizes the expected reward starting in $s$
\begin{align}
\pi = \arg \max_a Q (s,a).
\label{Eq:Q:pi}
\end{align}
If we prefer to work with cost $c(s,a)=-r(s,a)$, we can replace $r(s,a) $ with $c(s,a)$ in \eqref{Eq:Q} and define the policy as $\pi = \arg \min_a Q (s,a)$. 

An important observation is that \eqref{Eq:Q} is  actually a Bellman equation: The quality function \eqref{Eq:Q} of the current state-action pair $(s,a)$ is the immediate reward plus the quality of the next state-action pair $(s^{\prime}, \pi(s^{\prime}))$.

Finding the policy in \eqref{Eq:Q:pi} needs further consideration. To find the policy in each action, we need to solve an optimization problem; i.e. select the action $a$ to maximize $Q$. Since we have two possible scenarios where the action space can be discrete or continuous, we need to define the $Q$ function for each case properly so that it is possible to optimize the $Q$ function without appealing to advanced optimization techniques. From here on, we treat discrete and continuous action spaces differently.

\subsubsection{Discrete action space}
\label{Subsubsec:Q:Q:Discrete}
When there is a finite number of $n_a$ actions, we consider a network which takes the state $s$ as the input and generates $n_a$ outputs. Each output is $Q(s,a)$ for all $a \in \mathcal{A}$ and $Q(s,a)$ is obtained by indexing into the output vector \lstinline[language=Python]!network(state)!. The policy $\pi$ is the index which the output of the network is maximized.

\begin{figure}	
	\centering	
	\includegraphics[scale=0.4, trim=0cm 0cm 0cm 0cm,clip]{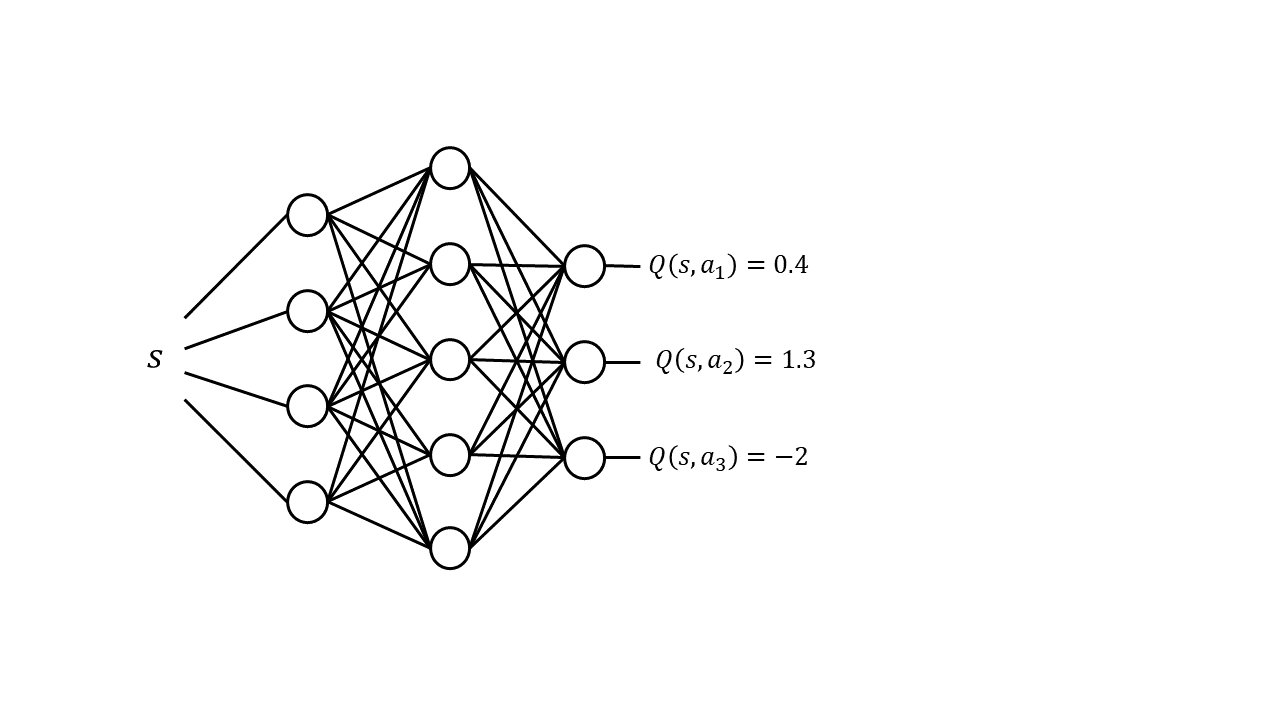}
	\caption{An example of network producing $Q(s,a)$ for all $a \in\{a_1,a_2,a_3\}$}
	\label{Fig:Q:discrete}
\end{figure}

For example, consider the network in Fig. \ref{Fig:Q:discrete}. This network takes the state $s$ as the input and generates $Q(s,a)$ for all possible actions $a \in \{a_1,\:a_2,\:a_3\}$. The policy for the state $s$ in this example is the index which the output of the network is maximized; i.e. $a_2$.

\begin{tcolorbox}[left skip=0.5cm, before skip=0.5cm,
	breakable,arc=8pt,borderline={0pt}{0pt}{white},boxrule=0mm,fontupper=\small]
	\textit{\textbf{Defining $Q$ function and policy in discrete action space case}}
	we consider a network which takes the state as the input and generates $n_a$ outputs.

	\begin{lstlisting}[language=Python]
	network = keras.Sequential([
	keras.layers.Dense(30, input_dim=n_s, activation='relu'),
	keras.layers.Dense(30, activation='relu'),
	keras.layers.Dense(30, activation='relu'),
	keras.layers.Dense(n_a)])
	\end{lstlisting}
	
	In the above code, we build the network. The network takes a state of dimension $n_s$ as the input and uses it in a fully connected layer with 30 neurons, with the activation function as relu, followed by two layers each with 30 neurons and with the activation function as relu. Then, we have the last layer which has $n_a$ number of outputs. The parameters in the networks are biases and weights in the layers. 
	
	Using the network which we just defined, we can define the policy as the argument that maximizes the $Q$ function
	\begin{lstlisting}[language=Python]
	policy = np.argmax(network(state))
	\end{lstlisting}
\end{tcolorbox}

\subsubsection{Continuous action space}
\label{Subsubsec:Q:Q:Continuous}
When the action space is continuous, we cannot follow the same lines as the discrete action space case because simply we have an infinite number of actions. In this case, the $Q$ function is built by a network which takes the state $s$ and action $a$ as the input and generates a single value $Q(s,a)$ as the output. The policy in each state $s$ is given by $\arg_{a} \max Q(s,a)$. Since we are not interested (neither possible nor making sense) in solving an optimization problem in each state, we select a structure for the $Q$ function such that the optimization problem is carried out analytically. One possible structure for the $Q$ function is quadratic which is commonly used in linear quadratic control problem \cite{Adib2021Linear}

\begin{align}
Q(s,a) =\begin{bmatrix}
s^{\dagger} & a^{\dagger}
\end{bmatrix}\begin{bmatrix}
g_{ss} & g_{sa}\\
g_{sa}^{\dagger} & g_{aa} \end{bmatrix}\begin{bmatrix}
s\\a \end{bmatrix}= z^{\dagger} G z
\label{Eq:Q:quadratic}
\end{align}
where $z=\begin{bmatrix}
s^{\dagger} & a^{\dagger}
\end{bmatrix}^\dagger $ and $G= \begin{bmatrix}
g_{ss} & g_{sa}\\
g_{sa}^\dagger & g_{aa} \end{bmatrix}$. The policy $\pi$ is obtained by mathematical maximization of the function $Q(s,a)$ with respect to $a$
\begin{align}
\pi(s) = -g_{aa}^{-1}g_{sa}^{\dagger} \: s.
\label{Eq:Q:quadratic:pi}
\end{align}

%

\subsection{Temporal difference learning}
\label{Subsec:Q:TD}
As the name implies, in a $Q$-learning algorithm, we build a (possibly deep) network and learn the $Q$-function. In the discrete action space case, the network takes the state $s$ as the input and generate $Q(s,a)$ for all $a\in \mathcal{A}$, see subsection \ref{Subsubsec:Q:Q:Discrete}. In the continuous action space, the network takes the state $a$ and action $a$ and generates $Q(s,a)$, see subsection \ref{Subsubsec:Q:Q:Continuous}. If this network represents the \textit{true} $Q$-function, then it satisfies the Bellman equation in \eqref{Eq:Q}. Before learning, however, the network does not represent the true $Q$ function. As a result, the Bellman equation \eqref{Eq:Q} is not satisfied and there is a \textit{temporal difference error} $e$

\begin{align}
e = r(s,a)+ \gamma\:\mathbf{E}[ Q(s^{\prime}, \pi(s^{\prime}))]-Q(s,a).
\label{Eq:Q:TD}
\end{align}
We learn the parameters in the network $Q$ to minimize the \textit{mean squared error (mse)} $\dfrac{1}{2} \: \sum_{t=1}^{T}e_{t}^2$. In the sequel, we show how to minimize the mean squared error in discrete and continuous action space cases.

\subsubsection{Discrete action space}
\label{Subsubsec:Q:TD:Discrete}
\begin{tcolorbox}[left skip=0.5cm, before skip=0.5cm,
	breakable,arc=8pt,borderline={0pt}{0pt}{white},boxrule=0mm,fontupper=\small]
	\textit{\textbf{Temporal difference learning in discrete action space case}}	
	To learn the parameters in the network, we define an mse cost for the network
	\begin{lstlisting}[language=Python]
	network.compile(loss='mean_squared_error')
	\end{lstlisting}
	After configuring the network, the last step is to feed the network with \textit{states, actions, rewards, next\_states,} and \textit{dones} and update the parameters of the network. Note that \textit{dones} is an array of Booleans with the same length as \textit{states}. The ith element in \textit{dones} is True if the ith state in \textit{states} is the last state in the episode (showing that the episode is ended) and False otherwise.
	
\begin{lstlisting}[language=Python]
eps_length = len(states)
states = np.vstack(states)
q_target = network(states).numpy
for i in range(eps_length):
	if dones[i]:
		q_target[i, actions[i]] = rewards[i]
	else:
		q_target[i, actions[i]] = rewards[i] + Gamma *
			tf.math.reduce_max(network(next_states[i])).numpy()
loss = network.train_on_batch(states, q_target)
\end{lstlisting}
We feed the network with \textit{states}. If the network correctly represents the $Q$ function, the output of the network would be the same as \textit{q\_target}. Usually it is not the case and there is an error (which is temporal difference error defined in \eqref{Eq:Q:TD}). As we have defined an mse cost function for the network, the parameters of the network is updated to minimize the mse error in the last line of the code.
\end{tcolorbox}

\subsubsection{Continuous action space}
\label{Subsubsec:Q:TD:Continuous}
For a quadratic $Q=z^{\dagger} G z$ function, the matrix $G$ is learned by \textit{Least Square Temporal Difference learning (LSTD)} \cite{Bradtke2004}
\begin{align}		
vecs(G)= ( \frac{1}{T}\sum_{t=1}^{T} \Psi_t (\Psi_t-\gamma \Psi_{t+1})^{\dagger} )^{-1} ( \frac{1}{T}\sum_{t=1}^{T} \Psi_t r_t ), 
\label{Eq:Q:quadratic:G}
\end{align}
where $\Psi_t=vecv(z_t),\:z_t=\begin{bmatrix}s_t^{\dagger} & a_t^{\dagger}\end{bmatrix}^{\dagger}$, see Table \ref{Table:notation} for the notations $vecs,\:vecv$.

\subsection{How to select action $a$? Exploration vs. Exploitation}
\label{Subsec:Q:EE}
You have probably heard about \textit{exploration vs. exploitation}. This concept is best described by this example. Suppose that you want to go to a restaurant in town. Exploration means that you select a random restaurant that you have not tried before. Exploitation means that you go to your favorite one. The good point with exploitation is that you like what you'll eat and the good point with exploration is that you might find something that you like more than your favorite. 

The same thing happens in RL. If the agent only sticks to exploitation, it can never improve its policy and it will get stuck in a local optimum forever. On the other hand, if the agent only explores, it never uses what it has learned and only tries random things. It is important to balance the levels of exploration and exploitation. The simplest way of selecting $a$ to have both exploration and exploitation is described here for discrete and continuous action space.

\subsubsection{Discrete action space}
\label{Subsubsec:Q:EE:discrete}
When there is a finite number of actions, the action $a$ is selected as follows. We set a level $0<\epsilon<1$ (for example $\epsilon = 0.1$) and we select a random number $r\sim [0,\:1]$. If $r<\epsilon$, we explore by selecting a random action otherwise, we follow the policy by maximizing the $Q$ function 

\begin{align*}
a = \begin{cases}
\text{random action}\quad \text{if   } r \:<\: \epsilon,\\
\arg \max_a Q (s,a) \quad \text{Otherwise}.
\end{cases}
\end{align*}

\begin{tcolorbox}[left skip=0.5cm, before skip=0.5cm,
	breakable,arc=8pt,borderline={0pt}{0pt}{white},boxrule=0mm,fontupper=\small]
	\textit{\textbf{Selecting action $a$ in discrete action space case}}
The following lines generate action $a$ with the exploration rate epsilon	
\begin{lstlisting}[language=Python]
if np.random.random() <= epsilon:
	selected_action = env.action_space.sample()
else:
	selected_action = np.argmax(network(state))
\end{lstlisting}	
where epsilon $\in [0,\:1]$. Note that smaller epsilon, less exploration. In the above lines, we generate a random number and if this number is less than epsilon, we select a random action; otherwise, we select the action according to the policy.
\end{tcolorbox}

\subsubsection{Continuous action space}
\label{Subsubsec:Q:EE:Continuous}
When the action space is continuous, the action $a$ is selected as the optimal policy plus some randomness. Let $r \sim \mathcal{N}(0,\sigma^2)$

\begin{align}
a = \arg \max_a Q (s,a) + r.
\label{Eq:Q:a}
\end{align}
\begin{tcolorbox}[left skip=0.5cm, before skip=0.5cm,
	breakable,arc=8pt,borderline={0pt}{0pt}{white},boxrule=0mm,fontupper=\small]
	\textit{\textbf{Selecting action $a$ in continuous action space case}}
When the $Q$ function is quadratic as \eqref{Eq:Q:quadratic} and the policy is given by \eqref{Eq:Q:pi}, a random action $a$ is selected as 
\begin{lstlisting}[language=Python]
a = -g_aa^{-1} @ g_sa.T @ state 
+ stddev * np.random.randn(n_a)
\end{lstlisting}
Note that smaller stddev, less exploration. (The symbol @ represent matrix multiplication.)
\end{tcolorbox}

\subsection{$Q$-learning as an algorithm}
\label{Subsec:Q:algorithm}
First, we build/select a network to represent $Q(s,a)$. See Subsection \ref{Subsec:Q}. Then, we iteratively improve the network. In each iteration of the algorithm, we do the following:

\begin{enumerate}
	\item  We sample a trajectory from the environment to collect data for $Q$-learning by following these steps:
	\begin{enumerate}
		\item We initialize empty histories for \textit{states=[], actions=[], rewards=[], next\_states=[], dones=[]}.
		
		\item We observe the state $s$ and select the action $a$ according to Subsection \ref{Subsec:Q:EE}.
		
		\item We derive the environment using $a$ and observe the reward $r$ and the next state $s^{\prime}$, and the Boolean $done$ (which is `True' if the episode has ended and `False' otherwise).
		
		\item We add $s,\:a,\:r,\:s^{\prime},\:done$ to the history batch \textit{states, actions, rewards, next\_states, dones}.
		
		\item We continue from 1.(b). until the episode ends.
	\end{enumerate}

	\item We use \textit{states, actions, rewards, next\_states, dones} to optimize the parameters of the network, see Subsection \ref{Subsec:Q:TD}. 
\end{enumerate}

\subsection{Improving $Q$-learning: Replay $Q$-learning}
We can improve the performance of $Q$-learning by some simple adjustments. The approach is called replay $Q$-learning and it has two additional components in comparison with the $Q$-learning.

\noindent
\textbf{Memory:} We build a memory to save data points through time. Each data point contains state $s$, action $a$, reward $r$, next\_state $s^{\prime}$, and the Boolean $done$ which shows if the episode ended. We save all the data sequentially. When the memory is full, the oldest data is discarded and the new data is added.

\noindent
\textbf{Replay:} For learning, instead of using the data from the latest episode, we sample the memory batch. This way we have more diverge and independent data to learn and it helps us to learn better.

\subsection{Replay $Q$-learning as an algorithm}
First, we build a network to represent $Q(s,a)$, see Subsection \ref{Subsec:Q:TD} and initiate an empty memory=[]. Then, we iteratively improve the network. In each iteration of the algorithm, we do the following:

\begin{enumerate}
	\item  We sample a trajectory from the environment to collect data for replay $Q$-learning by following these steps:
	\begin{enumerate}
		\item We observe the state $s$ and select the action $a$ according to Subsection \ref{Subsec:Q:EE}.
		
		\item We derive the environment using $a$, observe the reward $r$, the next state $s^{\prime}$ and the Boolean $done$.
		
		\item  We add $s,\:a,\:r,\:s^{\prime},\:done$ to \textit{memory}.	
		
		\item We continue from 1.(a). until the episode ends.
	\end{enumerate}
	
	\item We improve the $Q$ network
	\begin{enumerate}
		\item We sample a batch from \textit{memory}. Let \textit{states, actions, rewards, next\_states, dones} denote the sampled batch.
		
		\item We supply\textit{states, actions, rewards, next\_states, dones} to the network and optimize the parameters of the network. See Subsection \ref{Subsec:Q:TD}. One can see the difference between experience replay $Q$-learning and $Q$-learning here: In the experience replay $Q$ learning \textit{states, actions, rewards, next\_states, dones} are sampled from the memory but in the $Q$ learning, they are related to the latest episode.
	\end{enumerate}

\end{enumerate}

\section{Model Building, System Identification and Adaptive Control}
\subsection{Reinforcement Learning vs Traditional Approaches in Control	Theory: Adaptive Control}
\emph{Reinforcement Learning, RL, is about invoking actions (control) on the environment (the system) and  taking advantage of observations of the response to the actions to form better and better actions on the environment.} See Fig. \ref{Fig:RL:framework}.

The same words can also be used to define \emph{adaptive control} in standard control theory. But then typically another route is taken:
\begin{enumerate}
	\item See the environment or system as a mapping from  measurable inputs $u$ to measurable outputs $y$
	\item Build a mathematical model of the system (from $u$ to $y$) by some system identification technique. The procedure could be progressing in	time, so that at each time step $t$ a model $\theta(t)$  is available.
	\item Decide upon a desired goal for the control of  system, like that the output should follow a given reference signal  (that	could be a constant)
	\item Find a good control strategy for the goal, in case the system is described by the model $\theta^*$: $u(t)=h(\theta^*, y^t)$, where	$y^t$, denotes all outputs up to time $t$.
	\item Use the control policy $\pi: u(t)= h(\theta(t),y^t)$
\end{enumerate}

\begin{figure}	
	\centering	
	\includegraphics[scale=0.4, trim=0cm 0cm 0cm 0cm,clip]{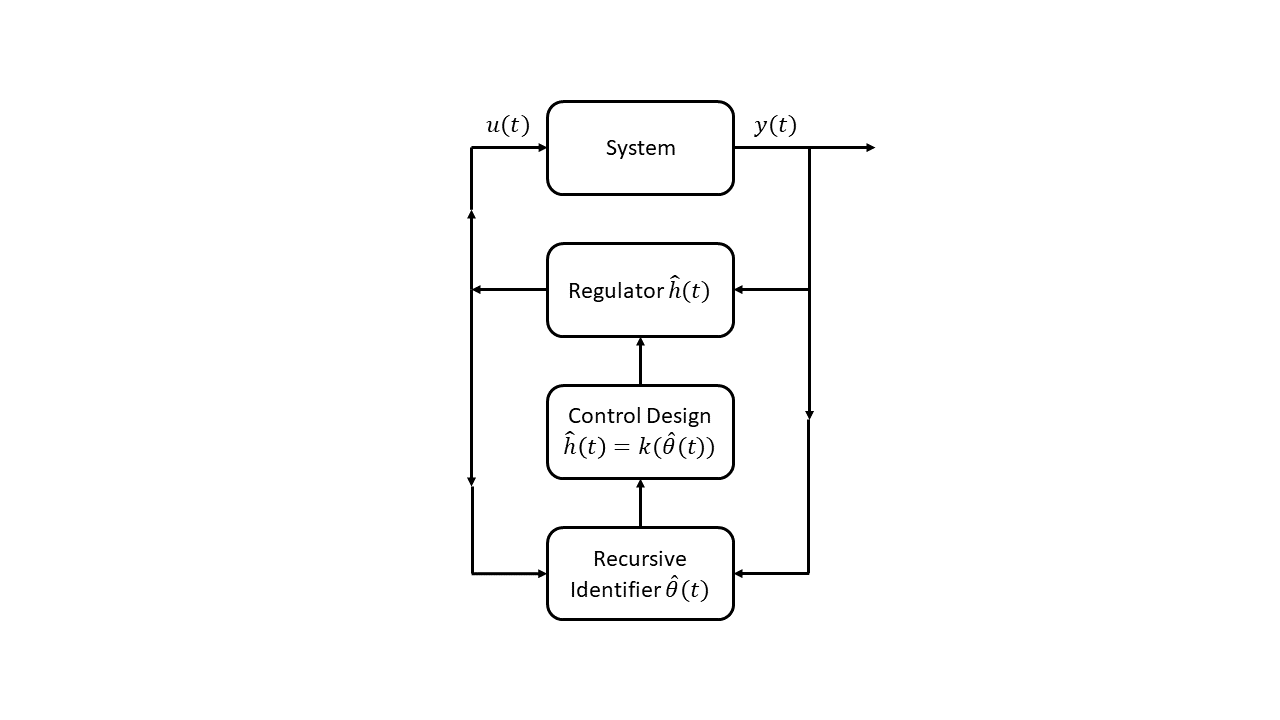}
	\caption{Model building approach}
	\label{Fig:MB:adaptive}
\end{figure}

See Fig. \ref{Fig:MB:adaptive}.

\subsection{System Identification}
System identification is about building mathematical  models of systems, based on observed inputs and outputs. It has three main ingredients:
\begin{itemize}
	\item The observed data $Z^t=\{y(t),y(t-1),\ldots,
	y(1),u(t-1),u(t-2),\ldots, y(t-N),u(t-N)\}$
	\item A \emph{model structure}, $\mathcal{M}:$ a parameterized set of
	candidate models $\mathcal{M}(\theta)$. Each model allows a
	prediction of the next output, based on earlier data: $\hat y(t|\theta)
	= g(t,\theta,Z^{t-1})$
	\item An \emph{identification method}, a mapping from $Z^t$ to $\mathcal{M}$ 
\end{itemize}

\begin{Ex}
A simple and common model structure is the ARX-model
\begin{align}
\label{eq:1}
y(t)+a_1y(t-1)+\ldots + a_n y(t-n)=b_1u(t-1)+\ldots +b_m u(t-m).
\end{align}
The natural predictor for this model is
\begin{align}
\begin{split}
\hat y(t|\theta)& = \varphi^T(t)\theta, \\
&\varphi^T(t) =[-y(t-1),\ldots -y(t-n), u(t-1), \ldots, u(t-m) ],\\
&\theta^T=[a_1, a_2 \ldots a_n, b_1, \ldots b_m].
\end{split}
\label{eq:arx}
\end{align}
The natural identification method is to minimize the Least Squares error between the measured outputs $y(t)$ and the model predicted output
$\hat y(t|\theta)$:
\begin{align}
\label{eq:3}
\hat \theta_N = \arg\min \sum_{t=1}^{N} \|y(t)-\hat y (t|\theta)\|^2.
\end{align}
Simple calculations give
\begin{align}
\label{eq:4}
\hat \theta_N &= D_N^{-1}f_N,\\
D_N& = \sum_{t=1}^{N} \varphi(t)\varphi^T(t) ;\quad f_N = \sum_{t=1}^{N} \varphi(t)y(t).
\end{align}
\end{Ex}

There are many other common model structures for system identification. Basically you can call a method (e.g. in the system identification toolbox in MATLAB) with your measured data and details for the structure and  obtain a model.
\begin{tcolorbox}[left skip=0.5cm, before skip=0.5cm,
	breakable,arc=8pt,borderline={0pt}{0pt}{white},boxrule=0mm,fontupper=\small]
	\textit{\textbf{Common model structures for system identification in the system identification toolbox in MATLAB }}		
	\begin{verbatim}
	m = arx(data,[na,nk,nb]) for the arx model above,
	m = ssest(data,modelorder) for a state space model
	m = tfest(data, numberofpoles) for a transfer function model
	\end{verbatim}
	\label{Shaded:adaptive}
\end{tcolorbox}

\subsection{Recursive System Identification}
The model can be calculated \emph{recursively} in time, so that it is updated any time new measurements become available. It is useful note that the least square estimate \eqref{eq:4} can be rearranged to be recalculated for each $t$:
\begin{align}
\label{eq:5}
\hat \theta(t) &= \hat \theta(t-1) + D_t^{-1} [y(t)-
\varphi^T(t)\hat \theta(t-1)] \varphi(t),\\
D_t& = D_{t-1} + \varphi(t)\varphi^T(t),
\end{align}
At time $t$ we thus only have to keep $  \hat \theta(t) , R_t$ in memory. This is the \emph{Recursive Least Squares, RLS} method.

Note that the updating difference $ [y(t)-\varphi^T(t)\hat \theta(t-1)] = y(t)-\hat y(t|\theta(t-1)$. The update is thus driven by the current model error.

Many variations of recursive model estimation can be developed for various model structure, but the RLS method is indeed the archetype for all recursive identification methods.

\subsection{Recursive Identification and Policy Gradient Methods in	RL}
There is an important conceptual, if not formal, connection between RLS and the Policy gradient method in Section \ref{Sec:PG}.

We can think of the \emph{reward} in system identification as to minimize the expected model error variance $J=E[\varepsilon(t,\theta)]^2$ where $\varepsilon(t,\theta)=y(t)-\hat y(t|\theta)$ (or maximize the negative value of it). The policy would correspond to the model parameters $\theta$. To maximize the reward wrt to the policies would mean to make adjustment guided by the gradient $\nabla J$. Now, for the ``identification reward'',  the gradient is (without expectation)
\begin{align}
\label{eq:6}
\nabla J &= 2 \varepsilon (-\psi) =2(y(t)-\hat y(t|\theta)\psi(t)),\\
\psi(t)& =\frac{d \hat y(t|\theta)}{d\theta}.
\end{align}
Note that for the ARX model (\ref{eq:arx}) $\psi(t) = \varphi(t)$ so the update in RLS is driven by the reward gradient. So in this way the recursive identification method can be interpreted as a  policy gradient method.

\appendix
\appendixpage
\section{RL on Cartpole Problem}
\label{App:Cartpole}
Cartpole is one of the classical control problems with discrete action space. In this section, we give a brief introduction to the cartpole problem and bring implementations of the PG, $Q$-learning and replay $Q$-learning for environments with discrete action spaces (like the cartpole environment). You can download the code for PG, $Q$-learning and replay $Q$-learning on the cartpole problem from the \href{https://github.com/FarnazAdib/Crash_course_on_RL/tree/master/cartpole}{folder `cartpole' in the Crash Course on RL.}

\subsection{Cartpole problem}
We consider cartpole which is a  classical toy problem in control. The cartpole system represents a simplified model of a harbor crane and it is simple enough to be solved in a couple of minutes with an ordinary PC.

\begin{figure}
	\centering
	\begin{subfigure}{.5\textwidth}
		\centering
		\includegraphics[width=.9\linewidth]{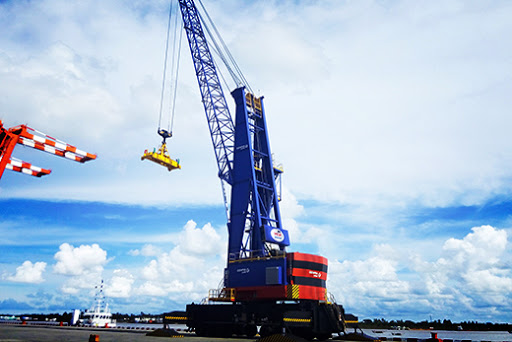}
		\caption{Photo credit @ http://rhm.rainbowco.com.cn/}
		\label{fig:sub1}
	\end{subfigure}%
	\begin{subfigure}{.5\textwidth}
		\centering
		\includegraphics[width=.9\linewidth]{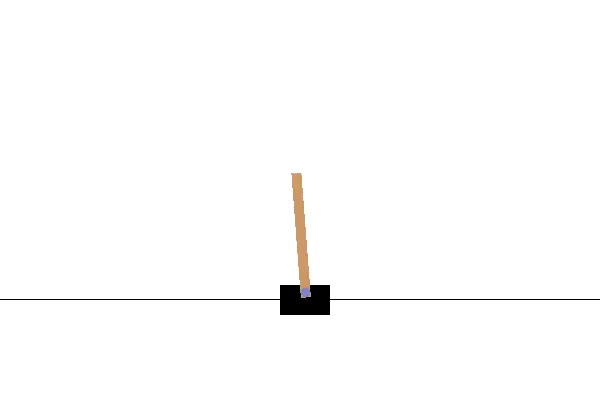}
		\caption{Photo credit @ https://gym.openai.com/}
		\label{fig:sub2}
	\end{subfigure}
	\caption{A harbor and a cartpole}
	\label{fig:test}
\end{figure}

\noindent
\textbf{Dynamics:} A pole is attached by an un-actuated joint to a cart. The cart is free to move along a frictionless track. The pole is free to move only in the vertical plane of the cart and track. The system is controlled by applying a force of +1 or -1 to the cart. The cartpole model has four state variables: 1- position of the cart on the track $x$, 2- angle of the pole with the vertical $\theta$, 3- cart velocity $\dot{x}$, and 4- rate of change of the angle $\dot{\theta}$.  The dynamics of cartpole system is governed by Newtonian laws and given in \cite{Barto83Neuronlike}.

We use the \href{https://gym.openai.com/envs/CartPole-v0/}{cartpole environment provided by OpenAI GYM which uses sampling time $0.02s$}. In this environment, the pole starts upright, and the goal is to prevent it from falling over. The episode ends when
\begin{itemize}
	\item the pole is more than 15 degrees from vertical or,
	\item the cart moves more than 2.4 units from the center or,
	\item the episode lasts for 200 steps.
\end{itemize}
The cartpole environments reveals a Boolean `done' which is always `False` unless the episode ends which becomes `True'.

\noindent
\textbf{Reward:} In each step, the cartpole environment releases an immediate reward $r_t$

\begin{align*}
r_t = \begin{cases}
1,\quad \text{if the pendulum is upright}\\
0,\quad \text{otherwise}
\end{cases}
\end{align*}
where ``upright'' means that $\vert x \vert < 2.4$ and $\vert \theta \vert < 12^{\circ}$.

\noindent
\textbf{Solvability criterion:} The CartPole-v0 defines \textit{solving} as getting average sum reward of 195.0 over 100 consecutive trials.

\noindent
\textbf{Why is cartpole an interesting setup in RL? }
\begin{itemize}
	\item The problem is small so it can be solved in a couple of minutes.
	\item The state space is continuous while the action space is discrete.
	\item This is a classical control problem. We love to study it!
\end{itemize}

\subsection{PG algorithm for the cartpole problem}
Here is a summary of PG algorithm for the cartpole problem (and it can be used for any other RL problem with discrete action space):

We build a (deep) network to represent the probability density function $\pi_\theta$= network(state), subsection \ref{Subsubsec:PG:pdf:Discrete} and assign a cross-entropy loss function, see subsection \ref{Subsubsec:PG:Gradient:Discrete}

\begin{lstlisting}[language=Python]
network = keras.Sequential([
	keras.layers.Dense(30, input_dim=n_s, activation='relu'),
	keras.layers.Dense(30, activation='relu'),
	keras.layers.Dense(n_a, activation='softmax')])
network.compile(loss='categorical_crossentropy')
\end{lstlisting}

Then, we iteratively improve the network. In each iteration of the algorithm, we do the following

\begin{enumerate}
	\item We sample a trajectory from the environment to collect data for PG by following these steps:
	\begin{enumerate}
		\item We initialize empty histories for \textit{states=[], actions=[], rewards=[]}.
		\item We observe the state $s$ and sample action $a$ from the policy pdf $\pi_{\theta}(s)$, see subsection \ref{Subsubsec:PG:pdf:Discrete}
\begin{lstlisting}[language=Python]
softmax_out = network(state)
a = np.random.choice(n_a, p=softmax_out.numpy()[0])
\end{lstlisting}	
		\item We derive the environment using $a$ and observe the reward $r$.
		\item We add $s,\:a,\:r$ to the history batch \textit{states, actions, rewards}.
		\item We continue from 1.(b) until the episode ends.
	\end{enumerate}
	\item We improve the policy by following these steps:
	\begin{enumerate}
		\item We calculate the reward to go and standardize it. 
		\item We optimize the policy, see subsection \ref{Subsubsec:PG:Gradient:Discrete}
\begin{lstlisting}[language=Python]
target_actions = tf.keras.utils.to_categorical
		(np.array(actions), n_a)
loss = network.train_on_batch
		(states, target_actions, 
		sample_weight=rewards_to_go)
\end{lstlisting}		
	\end{enumerate}
\end{enumerate}

\begin{tcolorbox}[left skip=0.5cm, before skip=0.5cm,
	breakable,arc=8pt,borderline={0pt}{0pt}{white},boxrule=0mm,fontupper=\small]
	\textit{\textbf{Here, we bring a simple class of implementing PG for an environment with discrete action space in python}}
\begin{lstlisting}[language=Python]
class PG:
	def __init__(self, hparams):
		self.hparams = hparams
		np.random.seed(hparams['Rand_Seed'])
		tf.random.set_seed(hparams['Rand_Seed'])
	
		# The policy network
		self.network = keras.Sequential([
			keras.layers.Dense(
				self.hparams['hidden_size'],
				input_dim=self.hparams['num_state'],
				activation='relu',
				kernel_initializer=
				keras.initializers.he_normal(), 
				dtype='float64'),
			keras.layers.Dense(
				self.hparams['hidden_size'],
				activation='relu',
				kernel_initializer=
				keras.initializers.he_normal(), 
				dtype='float64'),
			keras.layers.Dense(
				self.hparams['num_actions'],
				activation='softmax', 
				dtype='float64')])
				
		self.network.compile(
			loss='categorical_crossentropy',
			optimizer=keras.optimizers.Adam(
				epsilon=
				self.hparams['adam_eps'],
				learning_rate=
				self.hparams['learning_rate_adam']))

	def get_action(self, state, env):
	
		# Building the pdf for the given state
		softmax_out = self.network(state.reshape((1, -1)))
		
		# Sampling an action according to the pdf
		selected_action = np.random.choice(
			self.hparams['num_actions'],
			p=softmax_out.numpy()[0])
		return selected_action

	def update_network(self, states, actions, rewards):
		reward_sum = 0
		rewards_to_go = []
		for reward in rewards[::-1]:  # reverse buffer r
			reward_sum = reward 
				+ self.hparams['GAMMA'] * reward_sum
		rewards_to_go.append(reward_sum)
		rewards_to_go.reverse()
		rewards_to_go = np.array(rewards_to_go)
		# standardise the rewards
		rewards_to_go -= np.mean(rewards_to_go)
		rewards_to_go /= np.std(rewards_to_go)
		states = np.vstack(states)
		target_actions = tf.keras.utils.to_categorical(
			np.array(actions),
			self.hparams['num_actions'])
		loss = self.network.train_on_batch(
			states, target_actions,
			sample_weight=rewards_to_go)
		return loss

\end{lstlisting}
You can take a look at the integrated implementation of PG on the cartpole problem in
\href{https://github.com/FarnazAdib/Crash_course_on_RL/blob/master/pg_on_cartpole_notebook.ipynb}{Crash Course on RL}. 
\end{tcolorbox}

\subsection{$Q$-learning algorithm for the cartpole problem}
Here is a summary of $Q$-learning algorithm for the cartpole problem (and it can be used for any other RL problem with discrete action space):

We build a network to represent $Q(s,a)$, see subsection \ref{Subsubsec:Q:Q:Discrete} and assign a mean-square-error loss function, see subsection \ref{Subsubsec:Q:TD:Discrete}

\begin{lstlisting}[language=Python]
network = keras.Sequential([
keras.layers.Dense(30, input_dim=n_s, activation='relu'),
keras.layers.Dense(30, activation='relu'),
keras.layers.Dense(30, activation='relu'),
keras.layers.Dense(n_a)])

network.compile(loss='mean_squared_error', 
	optimizer=keras.optimizers.Adam())
\end{lstlisting}

Then, we iteratively improve the network. In each iteration of the algorithm, we do the following

\begin{enumerate}
	\item We sample a trajectory from the environment to collect data for $Q$-learning by following these steps:
	\begin{enumerate}
		\item We initialize empty histories for \textit{states=[], actions=[], rewards=[], next\_states=[], dones=[]}.
		\item We observe the state $s$ and select action $a$ according to (see subsection \ref{Subsubsec:Q:EE:discrete})
\begin{lstlisting}[language=Python]
if np.random.random() <= epsilon:
	selected_action = env.action_space.sample()
else:
	selected_action = np.argmax(network(state))
\end{lstlisting}	
		\item We derive the environment using $a$, observe the reward $r$, the next state $s^{\prime}$, and the Boolean $done$ (which is `True' if the episode has ended and `False' otherwise).
		\item We add $s,\:a,\:r,\:s^{\prime},\:done$ to the history batch \textit{states, actions, rewards, next\_states, dones}.
		\item We continue from 1.(b) until the episode ends.
	\end{enumerate}
	\item We supply \textit{states, actions, rewards, next\_states, dones} to the network and optimize the parameters of the network. See subsection \ref{Subsubsec:Q:TD:Discrete}
	
\begin{lstlisting}[language=Python]
eps_length = len(states)
states = np.vstack(states)
q_target = network(states).numpy

for i in range(eps_length):
	if dones[i]:
		q_target[i, actions[i]] = rewards[i]
	else:
		q_target[i, actions[i]] = rewards[i] + Gamma * 
		tf.math.reduce_max(network(next_states[i])).numpy()

loss = network.train_on_batch(states, q_target)
\end{lstlisting}		
\end{enumerate}

\begin{tcolorbox}[left skip=0.5cm, before skip=0.5cm,
	breakable,arc=8pt,borderline={0pt}{0pt}{white},boxrule=0mm,fontupper=\small]
	\textit{\textbf{Here, we bring a simple class of implementing $Q$-learning for an environment with discrete action space in python}}
\begin{lstlisting}[language=Python]
class Q:
	def __init__(self, hparams):
		self.hparams = hparams
		np.random.seed(hparams['Rand_Seed'])
		tf.random.set_seed(hparams['Rand_Seed'])
		self.epsilon = hparams['epsilon']

		# The Q network.
		self.network = keras.Sequential([
			keras.layers.Dense(
				self.hparams['hidden_size'], 
				input_dim=self.hparams['num_state'], 
				activation='relu',
				kernel_initializer=
				keras.initializers.he_normal(), 
				dtype='float64'),
			keras.layers.Dense(
				self.hparams['hidden_size'], 
				activation='relu',
				kernel_initializer=
				keras.initializers.he_normal(), 
				dtype='float64'),
			keras.layers.Dense(
				self.hparams['hidden_size'], 
				activation='relu',
				kernel_initializer=
				keras.initializers.he_normal(), 
				dtype='float64'),
			keras.layers.Dense(
				self.hparams['num_actions'], 
				dtype='float64')])

		# The cost function for the Q network
		self.network.compile(
			loss='mean_squared_error', 
			optimizer=keras.optimizers.Adam(
				epsilon=self.hparams['adam_eps'], 
				learning_rate=
				self.hparams['learning_rate_adam']))

	def get_action(self, state, env):
		state = self._process_state(state)
		
		if np.random.random() <= self.epsilon:
			# Exploration
			selected_action = env.action_space.sample()	
		else:
			# Exploitation
			selected_action = np.argmax(self.network(state))
	return selected_action

	def update_network(self, states, actions, rewards, next_states, 
			dones):
		eps_length = len(states)
		states = np.vstack(states)
		q_target = self.network(states).numpy()
		for i in range(eps_length):
			if dones[i]:
				q_target[i, actions[i]] = rewards[i]
			else:
				next_state = 
					self._process_state(next_states[i])
				q_target[i, actions[i]] = rewards[i] + 
					self.hparams['GAMMA'] * 
					tf.math.reduce_max(
					self.network(next_state)).numpy()
		loss = self.network.train_on_batch(states, q_target)
		return loss
	
	def _process_state(self, state):
		return state.reshape([1, self.hparams['num_state']])
	
\end{lstlisting}
You can take a look at the integrated implementation of $Q$-leaning on the cartpole problem in
\href{https://github.com/FarnazAdib/Crash_course_on_RL/blob/master/q_on_cartpole_notebook.ipynb}{Crash Course on RL}. 
\end{tcolorbox}

\subsection{Replay $Q$-learning algorithm for the cartpole problem}
Here is a summary of $Q$-learning algorithm for the cartpole problem (and it can be used for any other RL problem with discrete action space):

We build a network to represent $Q(s,a)$, see subsection \ref{Subsubsec:Q:Q:Discrete} and assign a mean-square-error loss function, see subsection \ref{Subsubsec:Q:TD:Discrete}

\begin{lstlisting}[language=Python]
network = keras.Sequential([
keras.layers.Dense(30, input_dim=n_s, activation='relu'),
keras.layers.Dense(30, activation='relu'),
keras.layers.Dense(30, activation='relu'),
keras.layers.Dense(n_a)])

network.compile(loss='mean_squared_error', 
optimizer=keras.optimizers.Adam())
\end{lstlisting}
We also initiate an empty \textit{memory=[]} for saving data.

Then, we iteratively improve the network. In each iteration of the algorithm, we do the following

\begin{enumerate}
	\item We sample a trajectory from the environment to collect data for replay $Q$-learning by following these steps:
	\begin{enumerate}
		\item We observe the state $s$ and select action $a$ according to (see subsection \ref{Subsubsec:Q:EE:discrete})
\begin{lstlisting}[language=Python]
if np.random.random() <= epsilon:
selected_action = env.action_space.sample()
else:
selected_action = np.argmax(network(state))
\end{lstlisting}	
		\item We derive the environment using $a$, observe the reward $r$, the next state $s^{\prime}$, and the Boolean $done$ (which is `True' if the episode has ended and `False' otherwise).
		\item We add $s,\:a,\:r,\:s^{\prime},\:done$ to \textit{memory}.
		\item We continue from 1.(a) until the episode ends.
	\end{enumerate}
	\item We improve the $Q$-network by following these steps:
	\begin{enumerate}
		\item We sample a batch from memory. Let \textit{states, actions, rewards, next\_states, dones} denote the sampled batch.
\begin{lstlisting}[language=Python]
batch = random.sample(memory, min(len(memory), batch_size))
states, actions, rewards, new_states, dones = 
	list(map(lambda i: [j[i] for j in batch], range(5)))
\end{lstlisting}	
		\item We supply \textit{states, actions, rewards, next\_states, dones} to the network and optimize the parameters of the network, see subsection \ref{Subsubsec:Q:TD:Discrete}	
\begin{lstlisting}[language=Python]
eps_length = len(states)
states = np.vstack(states)
q_target = network(states).numpy

for i in range(eps_length):
	if dones[i]:
		q_target[i, actions[i]] = rewards[i]
	else:
		q_target[i, actions[i]] = rewards[i] + Gamma * 
		tf.math.reduce_max(network(next_states[i])).numpy()

loss = network.train_on_batch(states, q_target)
\end{lstlisting}	
	\end{enumerate}

\end{enumerate}

\begin{tcolorbox}[left skip=0.5cm, before skip=0.5cm,
	breakable,arc=8pt,borderline={0pt}{0pt}{white},boxrule=0mm,fontupper=\small]
	\textit{\textbf{Here, we bring a simple class of implementing replay $Q$-learning for an environment with discrete action space in python.} This class has two additional functions remember and replay in comparison with the $Q$-learning class.}
\begin{lstlisting}[language=Python]
class Q:
	def __init__(self, hparams, epsilon_min=0.01, epsilon_decay=0.995):
		self.hparams = hparams
		np.random.seed(hparams['Rand_Seed'])
		tf.random.set_seed(hparams['Rand_Seed'])
		random.seed(hparams['Rand_Seed'])
		self.memory = deque(maxlen=100000)
		self.epsilon = hparams['epsilon']
		self.epsilon_min = epsilon_min
		self.epsilon_decay = epsilon_decay
		
		# The Q network
		self.network = keras.Sequential([
			keras.layers.Dense(
				self.hparams['hidden_size'], 
				input_dim=self.hparams['num_state'], 
				activation='relu',
				kernel_initializer=
				keras.initializers.he_normal(), 
				dtype='float64'),
			keras.layers.Dense(
				self.hparams['hidden_size'], 
				activation='relu',
				kernel_initializer=
				keras.initializers.he_normal(), 
				dtype='float64'),
			keras.layers.Dense(
				self.hparams['hidden_size'], 
				activation='relu',
				kernel_initializer=
				keras.initializers.he_normal(), 
				dtype='float64'),
			keras.layers.Dense(
				self.hparams['num_actions'], 
				dtype='float64')])
		
		# The cost function for the Q network
		self.network.compile(
			loss='mean_squared_error', 
			optimizer=keras.optimizers.Adam(
				epsilon=self.hparams['adam_eps'], 
				learning_rate=
				self.hparams['learning_rate_adam']))

	def get_action(self, state, env):
		state = self._process_state(state)
		
		if np.random.random() <= self.epsilon:
			# Exploration
			selected_action = env.action_space.sample()
		
		
		else:
			# Exploitation
			selected_action = np.argmax(self.network(state))
		return selected_action

	def remember(self, state, action, reward, next_state, done):
		# Adding data to the history batch
		self.memory.append(
			(state, action, reward, next_state, done))

	def replay(self, batch_size):
		# Sampling the history batch
		batch = random.sample(
			self.memory, min(len(self.memory), batch_size))
		states, actions, rewards, new_states, dones = list(
			map(lambda i: [j[i] for j in batch], range(5)))
		loss = self.update_network(
			states, actions, rewards, new_states, dones)
		
		# decreasing the exploration rate
		if self.epsilon > self.epsilon_min:
			self.epsilon *= self.epsilon_decay
		return loss

	def update_network(self, states, actions, rewards, next_states, 
		dones):
		eps_length = len(states)
		states = np.vstack(states)
		q_target = self.network(states).numpy()
		for i in range(eps_length):
			if dones[i]:
				q_target[i, actions[i]] = rewards[i]
			else:
				next_state = 
					self._process_state(next_states[i])
				q_target[i, actions[i]] = rewards[i] + 
					self.hparams['GAMMA'] * 
					tf.math.reduce_max(
					self.network(next_state)).numpy()
		loss = self.network.train_on_batch(states, q_target)
		return loss

	def _process_state(self, state):
		return state.reshape([1, self.hparams['num_state']])

\end{lstlisting}
You can take a look at the integrated implementation of $Q$-leaning on the cartpole problem in
\href{https://github.com/FarnazAdib/Crash_course_on_RL/blob/master/replay_q_on_cartpole_notebook.ipynb}{Crash Course on RL}. 
\end{tcolorbox}

\section{RL on Linear Quadratic Problem}
\label{App:LQ}
Linear Quadratic (LQ) problem is a classical control problem with continuous action space. In this section, we give a brief introduction to the LQ problem and bring implementations of the PG and $Q$-learning algorithms. We have not implemented replay $Q$-learning because the $Q$-learning algorithm performs superb on the LQ problem. You can download the code for PG and $Q$-learning on the LQ problem from the \href{https://github.com/FarnazAdib/Crash_course_on_RL/tree/master/lq}{folder `lq' in the Crash Course on RL.}

\subsection{Linear Quadratic problem}
Linear Quadratic (LQ) problem is a classical control problem where the dynamical system obeys linear dynamics and the cost function to be minimized is quadratic. The LQ problem has a celebrated closed-form solution and is an ideal benchmark for studying the RL algorithms because firstly, it is theoretically tractable and secondly, it is practical in various engineering domains. You can consider the Linear Quadratic problem as a simple example where you can derive the equations in this handout by some simple (but careful) hand-writing.

\noindent
\textbf{Dynamics}
We consider a linear Gaussian dynamical system

\begin{align}
s_{t+1}=A s_{t}+B u_{t}+ w_{t},
\label{Eq:LQ:Dynamics}
\end{align}
where $s_{t} \in \mathbb{R}^{n}$ and $u_{t} \in \mathbb{R}^{m}$ are the state and the control input vectors respectively. The vector $w_{t} \in \mathbb{R}^{n}$ denotes the process noise drawn i.i.d. from a Gaussian distribution $\mathcal{N}(\textbf{0}, W_{w})$. The linear system in \eqref{Eq:LQ:Dynamics} is an example of environment with continuous state and action spaces.

\noindent
\textbf{Cost}
In the LQ problem, it is common to define a quadratic running cost as
\begin{align}
c_{t}=s_{t}^{\dagger} Q s_{t}+u_{t}^{\dagger} R u_{t}
\label{Eq:LQ:running_cost}
\end{align}
where $Q \geq 0$ and $R>0$ are the state and the control weighting matrices respectively. It is enough to consider the reward as
\begin{align}
	r_t = - c_t.
\end{align}

\noindent
\textbf{Solvability criterion:} Define the value function associated with a policy $\pi$ as

\begin{align}
V(s_t)=\mathbf{E}[\sum_{k=t}^{+\infty} s_{k}^{\dagger} Q s_{k}+u_{k}^{\dagger} R u_{k}-\lambda) |s_{t}]=\mathbf{E}[c_{t}-\lambda + c_{t+1}- \lambda + ... | s_{t}]
\label{Eq:LQ:Value}
\end{align}
where $\lambda$ is the average cost associated with the policy $\pi$ 
\begin{align}
\lambda=\lim_{T \rightarrow \infty} \frac{1}{T} \sum_{t=1}^{T} c_{t} .
\label{Eq:LQ:Lam}
\end{align}
We aim to find a policy $\pi$ to minimize \eqref{Eq:LQ:Value}. 

A question may arise why we subtract $\lambda$ in \eqref{Eq:LQ:Value}. If we consider the value function as $V(s_t)=\mathbf{E}[\sum_{k=t}^{+\infty}(s_{kt}^{\dagger} Q s_{k}+u_{k}^{\dagger} R u_{k} )|s_{t}]$, the value function will be always infinite due to the process noise in \eqref{Eq:LQ:Dynamics} and it is not meaningful to minimize it. One possible mathematical modification is to consider minimizing the average cost \eqref{Eq:LQ:Lam}, which is finite. It has been shown that if $\pi$ minimizes \eqref{Eq:LQ:Value}, it also minimizes\eqref{Eq:LQ:Lam} \cite{bertsekas1995dynamic}.

\noindent
\textbf{Why is the LQ problem an interesting setup in RL? } But why do we consider to solve an LQ problem with RL when we can simply estimate the linear model?

\begin{itemize}
	\item The LQ problem has a celebrated closed-form solution. It is an ideal benchmark for studying the RL algorithms because we know the exact analytical solution so we can compare RL algorithms against the analytical solution and see how good they are.
	\item It is theoretically tractable.
	\item It is practical in various engineering domains.
\end{itemize}

\subsection{PG algorithm for the LQ problem}
For the LQ problem, we consider a Gaussian distribution with mean $\mu_{\theta}(s) =\theta \: s$ for the pdf of the policy, see subsection \ref{Subsubsec:PG:pdf:Continuous}. We iteratively improve the policy and in each iteration of the algorithm, we do the following

\begin{enumerate}
	\item We collect a number of batches. Each batch contains a sample a trajectory from the environment to collect data for PG by following these steps:
	\begin{enumerate}
		\item We initialize empty histories for \textit{states=[], actions=[], costs=[]}.
		\item We observe the state $s$ and sample action $a$ from the policy pdf $\pi_{\theta}(s)$, see subsection \ref{Subsubsec:PG:pdf:Continuous}
\begin{lstlisting}[language=Python]
a = theta s + sigma * np.random.randn(n_a)
\end{lstlisting}
		Note that $n_a$ is the dimension of the input in the continuous action space case, see Table \ref{Table:notation}.
		\item We derive the environment using $a$ and observe the cost $c$.
		\item We add $s,\:a,\:c$ to the history batch \textit{states, actions, costs}.
		\item We continue from 1.(b) until the episode ends.
	\end{enumerate}	
	\item We improve the policy by following these steps
	\begin{enumerate}
		\item We calculate the total reward \eqref{Eq:PG:undiscounted_total_reward} and standardize it. 
		\item We calculate the gradient from \eqref{Eq:PG:gradient:continuous:linear}, see subsection \ref{Subsubsec:PG:Gradient:Continuous}, which is
		\begin{align*}\nabla_{\theta} J = \dfrac{1}{\sigma^2 |\mathcal{D}|}\sum_{\tau \in \mathcal{D}} \sum_{t=1}^{T}(a_t-\theta \: s_t)s^{\dagger} (R(T)-b)
		\end{align*}
		where $b$ is a baseline.
		\item We update the parameter $\theta$ by a gradient descent algorithm.
	\end{enumerate}
\end{enumerate}

\begin{tcolorbox}[left skip=0.5cm, before skip=0.5cm,
	breakable,arc=8pt,borderline={0pt}{0pt}{white},boxrule=0mm,fontupper=\small]
	\textit{\textbf{Here, we bring a simple class of implementing PG for LQ problem (which has a continuous action space) in python.}}
\begin{lstlisting}[language=Python]
class PGRL:
	def __init__(self, sysdyn:Linear_Quadratic):

		self.dyn = sysdyn
		self.n, self.m = self.dyn.B.shape
		
	def safeK(self, K, safeguard):
		if np.isnan(K).any():
			K = safeguard * np.ones((self.m, self.n))
		return K
	
	
	def pg_linpolicy(self, K0, N, batch_size, T, 
		explore_mag=0.1, step_size=0.1, 
		beta1=0.9, beta2=0.999, epsilon=1.0e-8, 
		safeguard=10):
		'''
		
		:param K0: The initial controller gain
		:param N: Number of iterations
		:param batch_size: number of batches
		:param T: Trajectory length
		:param explore_mag: Magnitude of the noise to explore
		:param step_size: learning rate for Adam
		:param beta1: Adam related parameter
		:param beta2: Adam related parameter
		:param epsilon: Adam related parameter
		:param safeguard: The maximum of K
		:return: The  gain K
		'''
	
		# Initialize the controller
		Lin_gain = LinK(copy.copy(K0))
		Lin_gain.make_sampling_on(explore_mag)
	
		# A heuristic baseline to decrease the variance
		baseline = 0.0
		
		# Initializing Adam optimizer
		adam = ADAM(self.m, self.n, 
			step_size=step_size, 
			beta1=beta1, beta2=beta2, 
			epsilon=epsilon)
		
		# start iteration
		for k in range(N):
		batch = np.zeros((self.m, self.n))
		reward = np.zeros(batch_size)
		
		# In each iteration, collect batches
		for j in range(batch_size):
		
		# Do one rollout
		states, actions, costs, _ = self.dyn.one_rollout(
			Lin_gain.sample_lin_policy, T)
		
		# Building the gradient of the loss with respect to gain
		actions_randomness = actions - Lin_gain.lin_policy(states)
		reward[j] = -np.sum(costs)/T
		batch += explore_mag**(-2) * 
			((reward[j] - baseline) / batch_size) * 
			actions_randomness.T @ states
		
		# Update the baseline when batches are collected
		baseline = np.mean(reward)
		
		# Update the policy using ADAM
		dK = adam.opt_onestep(batch)
		Lin_gain.K += dK
		return self.safeK(Lin_gain.K, safeguard)

\end{lstlisting}
You can take a look at the integrated implementation of PG on the LQ problem in
\href{https://github.com/FarnazAdib/Crash_course_on_RL/blob/master/pg_on_lq_notebook.ipynb}{Crash Course on RL}. 
\end{tcolorbox}

\subsection{$Q$-learning algorithm for the LQ problem}
Because the dynamics is linear \eqref{Eq:LQ:Dynamics}, we consider a quadratic $Q$ function in \eqref{Eq:Q:quadratic} \cite{Adib2021Linear}, see subsection \ref{Subsubsec:Q:Q:Continuous}
\begin{align*}
Q(s,a) =\begin{bmatrix}
s^{\dagger} & a^{\dagger}
\end{bmatrix}\begin{bmatrix}
g_{ss} & g_{sa}\\
g_{sa}^T & g_{aa} \end{bmatrix}\begin{bmatrix}
s\\a \end{bmatrix}= z^{\dagger} G z
\end{align*}
where $z=\begin{bmatrix}
s^{\dagger} & a^{\dagger}
\end{bmatrix}^\dagger $ and $G= \begin{bmatrix}
g_{ss} & g_{sa}\\
g_{sa}^\dagger & g_{aa} \end{bmatrix}$. Remember that in the $Q$-learning, the policy $\pi$ is obtained by optimizing the $Q$ function with respect to the state and is given by (see subsection \ref{Subsubsec:Q:Q:Continuous})
\begin{align*}
\pi(s) = -g_{aa}^{-1}g_{sa}^{\dagger} \: s = K s.
\end{align*}

We start right away by selecting a stabilizing policy (or equivalently initializing the $Q$-function such that the resulting policy is stabilizing). In each iteration of the algorithm, we do the following

\begin{enumerate}
	\item We sample a trajectory from the environment using the current policy to compute the average cost
	\begin{align*}
	\lambda=\lim_{T \rightarrow \infty} \frac{1}{T}  \sum_{t=1}^{T} c_{t} .
	\end{align*}
	\item We sample a trajectory from the environment to collect data for $Q$ learning by following these steps:
	\begin{enumerate}
		\item We initialize empty histories for \textit{states=[], actions=[], costs=[], next\_states=[]}.
		\item We observe the state $s$ and select the action $a$ according to (see subsection \ref{Subsubsec:Q:EE:Continuous})
\begin{lstlisting}[language=Python]
a = K @ state + stddev * np.random.randn(n_a)
\end{lstlisting}
		Note that the symbol @ represent matrix multiplication.
		\item We derive the environment using $a$ and observe the cost $c$ and the next state $s^{\prime}$.
		\item We add $s,\:a,\:c,\:s^{\prime}$ to the history batch \textit{states, actions, costs, next\_states}.
		\item We continue from 2.(b) until the episode ends. 
	\end{enumerate}
\item We estimate the matrix $G$ as (see subsection \ref{Subsubsec:Q:TD:Continuous})

\begin{align*}		
vecs(G)= ( \frac{1}{T}\sum_{t=1}^{T} \Psi_t (\Psi_t- \Psi_{t+1})^{\dagger} )^{-1} ( \frac{1}{T}\sum_{t=1}^{T} \Psi_t (c_t - \lambda) ), 
\end{align*}
where $z_t = [s_t^{\dagger},\:a_t^{\dagger}]^{\dagger}$,   $\Psi_t=vecv(z_k)$.

\item We update the policy by
\begin{align*}
\pi = -g_{aa}^{-1}g_{sa}^{\dagger} \: s = K \: s.
\end{align*}
\end{enumerate}

\begin{tcolorbox}[left skip=0.5cm, before skip=0.5cm,
	breakable,arc=8pt,borderline={0pt}{0pt}{white},boxrule=0mm,fontupper=\small]
	\textit{\textbf{Here, we bring a simple class of implementing $Q$-learning algorithm for LQ problem (which has a continuous action space) in python.}}
\begin{lstlisting}[language=Python]
class Q_learning:
	def __init__(self, sysdyn:Linear_Quadratic):
		self.dyn = sysdyn
		self.n, self.m = self.dyn.B.shape
		self.n_phi = int((self.n + self.m) * 
			(self.n + self.m + 1) / 2)
		self.P = np.zeros((self.n, self.n))
	
	def ql(self, K0, N, T, explore_mag=1.0):
	'''
	Q learning loop 
		:param K0: The initial policy gain
		:param N: Number of iterations
		:param T: Trajectory length
		:param explore_mag: The amount of randomness in Q learning
		:return: kernel P and the controller gain K
		'''
		self.K = K0
		for k in range(N):
		
		# If the controller is stable, do an iteration
		if self.dyn.is_stable(self.K):
	
			# Policy evaluaion
			G = self.q_evaluation(T, explore_mag)
			
			# Policy improvement
			self.K = self.q_improvement(G)
			P = GtoP(G, self.K)
		
		# If the controller is not stable,
		# return some unstable values for P and K
		else:
			P, self.K = self.unstable_P_and_K()
		
		return P, self.K
	
	def q_evaluation(self, T, explore_mag):
		# creating the linear policy and setting exploration rate
		Lin_gain = LinK(self.K)
		Lin_gain.make_sampling_on(explore_mag)
		
		# Do one rollout to compute the average cost
		_, _, c, _ = self.dyn.one_rollout(Lin_gain.lin_policy, T)
		Lam = np.sum(c)/T
		
		# Do one rollout to save data for Q learning
		states, actions, costs, next_states = 
			self.dyn.one_rollout(Lin_gain.sample_lin_policy, 
				T)
		
		# Making z and the next z
		z = np.concatenate((states, actions), axis=1)
		next_z = np.concatenate(
			(next_states, Lin_gain.lin_policy(next_states)),
			 axis=1)
		
		# estimating the Q parameter using instrumental variable
		x_iv = vecv(z) - vecv(next_z)
		y_iv = costs - Lam
		z_iv = vecv(z)
		q_vec = inst_variable(x_iv, y_iv, z_iv)
		return SquareMat(q_vec, self.n+self.m)
	
	def q_improvement(self, G):
		return - LA.inv(G[self.dyn.n:, self.dyn.n:]) @ 
			G[self.dyn.n:, 0:self.dyn.n]
	
	def unstable_P_and_K(self):
		return np.zeros((self.n, self.n)), 
			np.zeros((self.m, self.n))
\end{lstlisting}
You can take a look at the integrated implementation of $Q$-learning on the LQ problem in
\href{https://github.com/FarnazAdib/Crash_course_on_RL/blob/master/q_on_lq_notebook.ipynb}{Crash Course on RL}. 
\end{tcolorbox}

\section*{Acknowledgement}
We thank Fredrik Ljungberg for providing us much useful feedback on the repository \href{https://github.com/FarnazAdib/Crash_course_on_RL/blob/master/q_on_lq_notebook.ipynb}{``A Crash Course on RL"}. Farnaz Adib Yaghmaie is supported by the Vinnova Competence Center LINK-SIC, the Wallenberg Artificial Intelligence, Autonomous Systems and Software Program (WASP), and Center for Industrial Information Technology (CENIIT).

\bibliographystyle{IEEEtran}
\bibliography{IEEEabrv,RL}
\end{document}